\PassOptionsToPackage{colorlinks}{hyperref}  
\documentclass{article}
\usepackage[utf8]{inputenc}
\usepackage[a4paper, margin=1in]{geometry}

\usepackage{natbib}
\usepackage{fancyhdr, fullpage, parskip,amsfonts}
\usepackage{times,latexsym}
\usepackage[T1]{fontenc}

\usepackage{hyperref}
\usepackage{url}            
\usepackage{booktabs}       
\usepackage{amsfonts}       
\usepackage{nicefrac}       
\usepackage{microtype}      
\usepackage{xcolor}         
\usepackage{graphicx} 
\usepackage{wrapfig,lipsum,booktabs}
\usepackage{tcolorbox}
\usepackage{geometry}
\usepackage{amssymb}
\usepackage{amsmath}
\usepackage{multirow}
\usepackage{booktabs}
\usepackage{enumitem}
\usepackage{adjustbox}
\usepackage{makecell}
\usepackage{pifont}
\usepackage{bbm}
\usepackage{xspace}
\usepackage{subcaption}     
\usepackage{algorithm} 
\usepackage{algpseudocode}

\newtheorem{proposition}{Proposition}[section]

\newtheorem{definition}{Definition}[section]
\newcommand{\customfootnotetext}[2]{{%
      \renewcommand{\thefootnote}{#1}%
      \footnotetext[0]{#2}}}%

\newenvironment{myquotation}{\setlength{\leftmargini}{0em}\quotation}{\endquotation}

\newcommand{\projfull}{{Refined Graph-based RAG}\xspace}
\newcommand{\proj}{\texttt{ReG}\xspace}
\newcommand{\projj}{\texttt{ReG}}
\newcommand{\cp}{\mathcal{P}}

\newcommand{\cd}{\mathcal{D}}
\newcommand{\ca}{\mathcal{A}}
\newcommand{\cH}{\mathcal{H}}

\newcommand{\cs}{\mathcal{S}}

\newcommand{\ce}{\mathcal{E}}

\newcommand{\be}{\mathbb{E}}
\newcommand{\cR}{\mathcal{R}}
\newcommand{\cE}{\mathcal{E}}
\newcommand{\cg}{\mathcal{G}}
\newcommand{\ct}{\mathcal{G}}
\newcommand{\wt}{\widehat{\mathcal{G}}}

\definecolor{mydarkblue}{rgb}{0,0.08,0.45}

\begin{document}

\title{Weak-to-Strong GraphRAG: Aligning Weak Retrievers with Large Language Models for Graph-based Retrieval Augmented Generation}

\hypersetup{
    colorlinks=true,
    linkcolor=red,
    citecolor=mydarkblue,
    urlcolor=blue
}

\author{
  Deyu Zou$^{*\dag}$
  \and
  Yongqiang Chen$^{*\dag}$
  \and
  Mufei Li$^{\ddag}$
  \and
  Siqi Miao$^{\ddag}$
  \and
  Chenxi Liu${^\S}$
  \and
  Bo Han$^\S$
  \and
  James Cheng$^{\diamond\dag}$
  \and
  Pan Li$^{\diamond\ddag}$
}

\date{
  $^\dag$ The Chinese University of Hong Kong \\ $^\ddag$ Georgia Institute of Technology \\
  $^\S$ Hong Kong Baptist University \\
}

\maketitle
\customfootnotetext{$*/\diamond$}{Equal first-author / senior contribution.}

\begin{abstract}
\noindent
  Graph-based retrieval-augmented generation (RAG) enables large language models (LLMs) to ground responses with structured external knowledge from up-to-date knowledge graphs (KGs) and reduce hallucinations.
  However, LLMs often rely on a \textit{weak retriever} in graph-based RAG:
  I) Due to the lack of ground truth, the retriever is often trained on \textit{weak supervision}, which often introduces \textit{spurious signals} to the LLMs.
  II) Due to the abstraction of graph data, the retrieved knowledge is often presented in \textit{unorganized} forms.
  To mitigate the issue, we present \projfull (\proj) to align weak retrievers to LLMs for graph-based RAG.
  Specifically, \proj incorporates LLM feedback to get rid of spurious signals and improve the quality of the supervision. Meanwhile, \proj introduces a structure-aware reorganization module to refactor the retrieval results into logically coherent evidence chains.
  Experiments on prominent benchmarks demonstrate that \proj significantly and consistently brings improvements across different LLM backbones by up to 10\%.
  The improved supervision quality enables \proj to match the state-of-the-art performance with 5\% training data and to transfer to out-of-distribution KGs.
  Notably, when adopted to reasoning-based LLMs, \proj reduces the reasoning token cost by up to 30\% and improves the performance by up to 4\%.
\end{abstract}

\section{Introduction}
Large language models (LLMs) have achieved remarkable success and continue to advance~\citep{bang2023multitask,raiaan2024review,yang2024harnessing}. In particular, long-context LLMs~\citep{liu2025comprehensive,huang2023advancing,zhao2023length} extend context windows to millions of tokens~\citep{team2024gemini}, and large reasoning models (LRMs) 
like OpenAI o3~\citep{openai2025o3mini} and DeepSeek R1~\citep{guo2025deepseek} further push the frontier on extremely complex tasks. 
However, even frontier LLMs fall short in processing extremely lengthy or noisy input contexts: long-context LLMs often fail to effectively extract the useful information from the long context~\citep{hsieh2404ruler,liu2024lost},
while LRMs tend to incur excessive reasoning costs on irrelevant or redundant content~\citep{chen2024not,sui2025stop,feng2025efficient}. Therefore, it is essential to retrieve the desired information for LLMs, especially when handling complicated multi-hop reasoning tasks~\citep{han2024retrieval}.

Retrieval-augmented generation (RAG) has emerged as a powerful paradigm for enhancing LLMs~\citep{borgeaud2022improving,gao2023retrieval,chen2024benchmarking,fan2024survey}. By retrieving query-relevant information from external knowledge sources, RAG reduces LLM hallucinations~\citep{ji2023survey,huang2025survey,liu2024survey} and improves   the factuality~\citep{dhingra2022time,kasai2023realtime} and specificity~\citep{li2023chatgpt} of LLM responses.
Recently, graph-based RAG has extended text-based RAG to further enhance retrieval effectiveness in capturing the interconnections between facts. 
By retrieving from graph-structured textual and document-level knowledge in a knowledge graph (KG), graph-based RAG provides more structure-aware, interpretable, and compositional information compared to text-based retrieval, particularly effective for handling complex multi-hop reasoning tasks~\citep{chein2008graph,robinson2015graph,edge2024local,peng2024graph,han2024retrieval}.

\begin{figure*}[t]
    \vspace{-0.15in}
     \centering
\includegraphics[trim=160 15 0 15,clip,width=\linewidth]{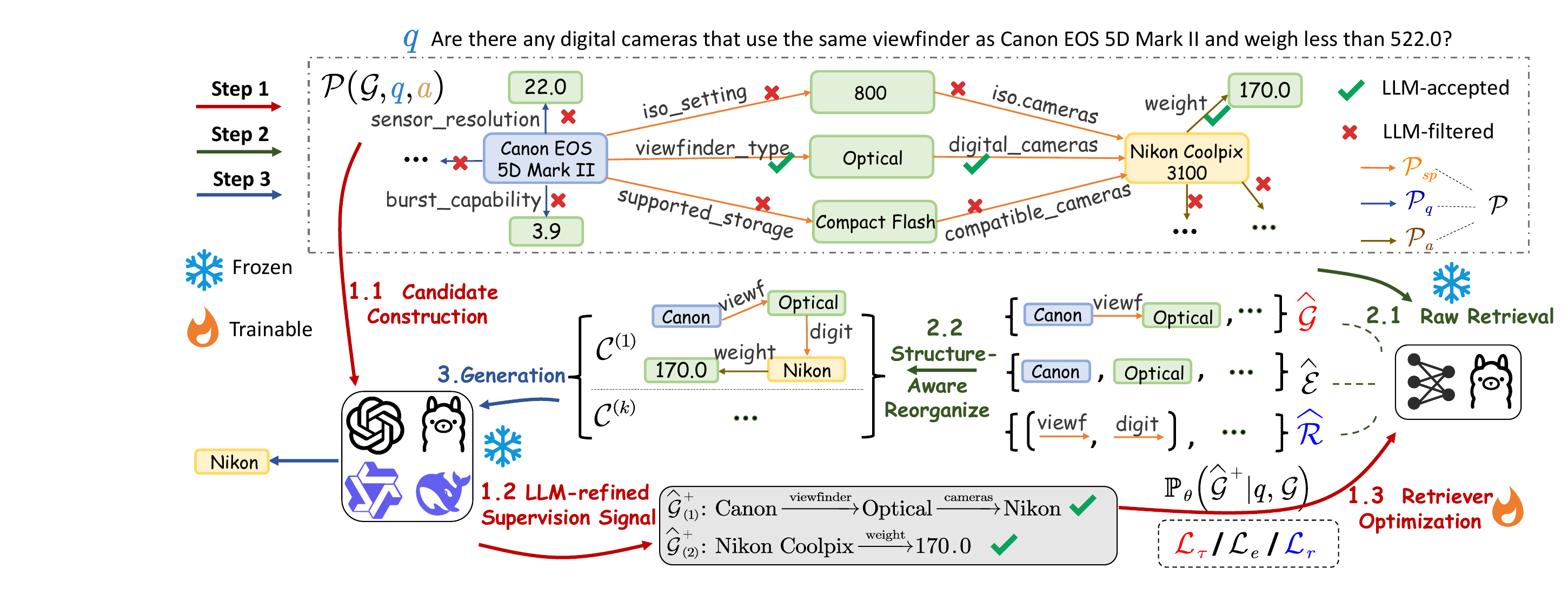}

     \caption{
     Overall framework of \proj.
Given a KG $\cg$ with a query-answer pair $(q,a)$, \proj first constructs a candidate path pool $\cp$ to cover diverse candidate reasoning paths. Then, \proj uses LLM to select high-quality $\wt^+$ for training retrievers. During inference, retrieved items are reorganized into logic-consistent chains to better align with the needs for LLM reasoning.
}
    \label{fig:pipeline}
    \vspace{-.2in}
\end{figure*}

Despite its promise, graph-based RAG relies on \textit{weak retrievers} that are often \textit{misaligned} with LLMs:
\textbf{(I) Weak supervision}.
Unlike text-based RAG, there is no general-purpose retriever tailored for structural information~\citep{luo2025gfm}, and the ground truth for graph-based RAG is typically unavailable. Consequently, graph-based retrievers often require training on specific datasets with \textit{heuristic-based weak supervision signals}~\citep{zhang2022subgraph,luo2023reasoning}. 
However, such supervision often misses key supporting evidence or includes spurious connections unrelated to the underlying  reasoning logic. This issue becomes  particularly pronounced  for multi-hop KG queries, where errors or omissions in key intermediate reasoning steps in the supervision will severely degrade retriever performance.
\textbf{(II) Representation misalignment}. The raw retrieved subgraph is often returned as an unstructured set of triples or entities~\citep{mavromatis2024gnn,li2024simple},   lacking logical coherence and contextual order. However,  LLMs are sensitive to the structure and ordering of input~\citep{chen2024premise,guo2025llms}, and such disorganized representations can hinder LLM reasoning and suppress the utility of correctly retrieved facts, which further increases the complexity of graph-based RAG. 
These dual issues create a fundamental misalignment between the capabilities of graph retrievers and the reasoning needs of LLMs, raising a critical and challenging research question: 
\begin{myquotation}\centering
    \textit{How can one effectively align weak graph retrievers with LLMs in graph-based RAG?}
\end{myquotation}
To tackle this core challenge  of aligning graph retrievers with LLMs, we present \projfull (\proj). It first incorporates the rich knowledge of LLMs to 
refine and align the weak supervision in graph-based RAG.
Essentially, we show that graph-based RAG can be  formulated as a black-box combinatorial search over the KG $\cg$: given a query $q$, the goal is to identify a minimal sufficient subgraph $\wt^* \subseteq \cg$ for an LLM to answer $q$ correctly. Here, the LLM serves as a black-box evaluator that assesses the utility of retrieved subgraphs.
Under this formulation, we show that solving the original black-box optimization problem is computationally intractable and thus not feasible under realistic LLM usage budgets. 
Therefore, \proj incorporates the LLMs in a simple yet effective way: using LLMs to select better reasoning chains among the candidate chains extracted from KG. The resulting  improved supervision signal improves the identification of the optimal subgraph in a cost-efficient manner. Moreover,
to further  align the retrieval results with LLMs, \proj reorganizes the retrieved content into logic-preserving chains, which is simple but significantly mitigates distraction and inefficiency during LLM reasoning.

Extensive experiments demonstrate that  \proj achieves state-of-the-art results on prominent multi-hop knowledge graph question answering (KGQA) benchmarks.
Notably, it yields retrievers with stronger zero-shot generalizability to out-of-distribution (OOD) KGs, mitigating the weakness of lacking foundation models in graph-based RAG.
The improved supervision enables \proj to match the state-of-the-art performance with 5\% training data.
\proj also brings measurable gains while reducing token costs by up to 30\%  when paired with the frontier LRMs. 
These results highlight the effectiveness, data-efficiency, and versatility of \proj as a generally applicable framework that enhances reasoning quality even when used with the most capable frontier LLMs.  

\section{Preliminaries and Related Works}
\textbf{Preliminaries.} In this work, we focus on \textbf{knowledge graph question answering} (KGQA), a central task in graph-based RAG: 
Given a natural language query $q$ and a KG $\cg$, the LLM answers the query by reasoning over the retrieved subgraph $\widehat{\cg} \subseteq \cg$, which ideally captures all critical supporting evidence while minimizing distracting noise.
A \textbf{knowledge graph} is a structured representation of factual knowledge, typically formulated as a collection of triples $\cg = \{(h, r, t) \mid h, t \in \cE, r \in \cR\}$, where $\cE$ and $\cR$ denote the sets of entities and relations, respectively. Each triple $\tau=(h, r, t)$ represents a directed relation $r$ from head entity $h$ to tail entity $t$.
Given the query $q$, we denote by $\mathcal{E}_q \subseteq \mathcal{E}$ the set of \textbf{query entities} extracted from $q$, and by $\mathcal{A}_q \subseteq \mathcal{E}$ the corresponding set of \textbf{answer entities}.
A \textbf{reasoning path} is defined as an ordered sequence of connected triples $P = (\tau_1, \dots, \tau_k)$, where each $\tau_i = (h_i, r_i, t_i) \in \cg$ satisfies $t_i = h_{i+1}$ for all $1 \leq i < k$.

\textbf{Graph-based Retrieval.}
Prior work has explored retrieving from existing KGs like WikiData~\citep{vrandevcic2014wikidata} or Freebase~\citep{bollacker2008freebase}, or augmenting text corpora with graph overlays~\citep{edge2024local,gutierrez2024hipporag,guo2024lightrag} to improve relevance modeling during retrieval. 
Due to the lack of an oracle in graph-based RAG, it often relies on heuristic priors or weak supervisions to guide the retrieval, which can be broadly categorized as:
\textit{Training-free} methods typically adopt graph-based heuristics (\textit{e.g.,} personalized PageRank)~\citep{gutierrez2024hipporag,gutierrez2025rag} or LLM-guided stepwise exploration over the graph~\citep{gu2022don,sun2023think,xiong2024interactive}.
However, graph algorithms often underperform as they struggle to combine semantic and structural information~\citep{luo2025gfm}, while LLM-guided traversal suffers from high computational cost and is prone to local decision biases~\citep{luo2025kbqa}.
\textit{Training-based} methods  train parametric  retrievers 
to recognize critical substructures~\citep{li2024simple,mavromatis2024gnn,he2024g}.
However, 
the training supervision is typically derived from noisy proxies such as query-answer shortest paths~\citep{zhang2022subgraph,luo2023reasoning}, which may omit critical intermediate evidence or introduce spurious connections unrelated to the reasoning logic~\citep{li2024simple}.
Differently, we aim to bridge the gap by integrating the rich knowledge of LLMs to improve the quality of weak supervision for retriever training.

\textbf{RAG with LLM Feedback.} Recent efforts in text-based RAG have explored using LLM feedback to directly optimize the retriever capability. 
\cite{shi2023replug} aligns retriever output distributions with the perplexity-reducing behavior of LMs.
\cite{li2024rag} iteratively tunes the LLM generator and retriever by aligning data preferences between both modules.
\cite{han2025gasketrag} train an intermediary module by collecting preference data from both the LLM and the retriever to enhance RAG performance. 
While effective for document-level retrieval, 
in graph-based RAG, the combinatorial explosion of subgraph candidates and their intricate dependencies pose unique challenges, making direct adaptation of these methods non-trivial.

\textbf{RAG with Post-retrieval Refinement.} In addition, to compensate for retrieval noise, prior work has also explored various post-hoc refinement strategies, including re-ranking retrieved contents~\citep{jin2024long}, filtering irrelevant spans~\citep{chen2025pathrag,wang2023learning,guo2025empowering}, or fine-tuning LLMs to tolerate noisy input~\citep{yoran2023making,yu2024rankrag,zhang2024raft,jin2024long}. While these methods improve robustness,  they largely assume the initial retrieval output to be fixed. Consequently, they hardly recover critical evidence missing from the initial retrieval stage, nor improve the capability of the upstream retriever itself.

\section{Drawbacks of Weak Retrievers in Graph-based RAG}\label{Formulations}

\subsection{Graph-based RAG as Black-box Combinatorial Optimization}\label{challenge}
We begin with a formal description of the problem to understand the influence of weak retrievers.
Essentially, the graph-based RAG can be considered as a black-box combinatorial optimization problem.
Given the query $q$ for a KG $\cg$ with $N$ triples, let $s(\cdot,q):\ct\mapsto[0,1]$ be an unknown scoring function assigning relevance scores to items $\tau\in\cg$, where $s(\tau,q)>0$ indicates $\tau$ is relevant to $q$.
The retriever aims to find the underlying \textit{unknown oracle set} $\wt^* \subseteq \ct$, defined as $\wt^* := \{\tau \in \ct \mid s(\tau,q) > 0\}$, and assumed to be sparse, \textit{i.e.}, 
$|\wt^*|\sim\mathcal{O}(1)\ll|\ct|$, which enables LLMs to answer the question correctly. Hence, LLMs can be considered as the black-box evaluator of the retrieved information (\textit{i.e.}, a subgraph $\wt$ of the KG).

\begin{definition}[Black-box LLM evaluator]\label{def-reward}
A black-box reward function $r(\cdot,q):2^{\ct}\mapsto\mathbb{R}$ evaluates how well a subset $\wt\subseteq\ct$ aligns with the oracle set $\wt^*$ by the following properties: 

i) \textit{Aggregation}. The reward aggregates relevance scores $s(\tau,q)>0$ for relevant items $\tau\in(\wt\cap\wt^*)$ and penalties $\delta(\tau,q)\geq0$ for noisy items $\tau\in(\wt\backslash\wt^*)$ via a black-box function $f$. Formally,
\begin{align}
    r(\wt, q) := f\left(\{s(\tau,q) \mid t \in (\wt\cap\wt^*)\}, \{\delta(\tau,q) \mid \tau\in(\wt\backslash\wt^*)\}\right),
\end{align}
where $\delta$ quantifies the LLM evaluator’s robustness to noise:  $\delta(\tau,q)=0$  means full robustness, while $\delta(\tau,q)>0$ imposes a penalty on $r(\wt,q)$.

ii) \textit{Exactness.} For any $\wt\subset\ct$, the reward reaches its maximum \textit{if and only if} $\wt=\wt^*$.
For uniform scores $s(\cdot,q)=s_0>0$ and $\delta(\cdot,q)=\delta_0>0$, a plausible instantiation of $r$ would be:
\begin{align}
    r(\wt,q):=\frac{|\wt\cap\wt^*|s_0-|\wt\backslash \wt^*|\delta_0}{|\wt^*|s_0}\,\,\in[-\frac{\delta_0}{s_0},1] \label{reward}
\end{align}
\end{definition}

Nevertheless, due to the lack of the oracle set $\wt^*$, existing methods resort to weak supervision signals $\wt^w$ derived from some heuristics, such as query-answer ($q$-$a$) shortest paths~\citep{zhang2022subgraph,luo2023reasoning}. While easy to extract, this approximation introduces a fundamental mismatch between $\wt^*$ and $\wt^w$, undermining retriever training in two key ways:
\textbf{(I) Incompleteness ($\wt^* \setminus \wt^w \ne \varnothing$).} The shortest-path supervision often omits essential reasoning components required to justify an answer. 
\textbf{(II) Spurious inclusion ($\wt^w \setminus \wt^* \ne \varnothing$).} Conversely, shortest paths may include semantically irrelevant or misleading information. 
The examples of these two cases can be seen in Appendix~\ref{app:eg-shortest}.
Together, these issues underscore the need for more faithful supervision signals that better approximate $\wt^*$.

\subsection{Computational Challenge in Iteratively Refined Graph-based RAG}\label{challenge:1}
To cope with the weak supervision, it is necessary to \textit{refine the retrieved contents iteratively}. 
In other words, the retriever needs to maximize $r$ under a strict budget of LLM evaluations $C\ll {|\ct|}$:
\begin{align}\label{eq:iterative_graphrag}
    \max\, r(\wt,q),\,\,\text{s.t.},\,\wt\in\{\wt^{(i)}\}_{i=0}^T,\,T\leq C
\end{align}
where $\{\wt^{(i)}\}_{i=0}^T$ is a sequence of subsets refined iteratively, initialized from $\wt^{(0)}=\oslash$ and updated via  $\wt^{(i+1)}=\textsc{Alg}\left(\left\{\wt^{(j)},r(\wt^{(j)},q)\right\}_{j=0}^i\right)$, and $\textsc{Alg}(\cdot)$ denotes a certain iterative graph-based RAG algorithm. However, as shown in the following proposition, solving for Eq.~\ref{eq:iterative_graphrag} is intractable.
\begin{proposition}\label{prop:general}
    For any algorithm interacting with $r(\cdot,q)$ in Eq.~\ref{reward} and $\ct$, after $T$ rounds, achieving 
    \begin{align}
        \mathbb{P}\left(\exists i\in[T]: r\left(\wt^{(i)}, q\right) =1\right) \geq 1-\varepsilon
    \end{align}
    with $\varepsilon\in(0,1)$ requires
    \begin{align}
        T\geq \Omega\left(\frac{(1-\varepsilon)N}{\log N}\right).
    \end{align}
    {Additionally, if $|\wt^{(i)}|,i\in[T]$ is fixed as a constant, then $T\geq \Omega((1-\varepsilon)N)$.}
\end{proposition}
See Appendix~\ref{proof:main} for the proof. In practice, since $N$ is large, identifying the $\wt^*$ is essentially computationally intractable, which motivates us to consider more cost-efficient approaches.

\subsection{Representation Challenge in Retrieved Knowledge}\label{challenge:2}
Beyond the two key properties of 
$r(\cdot,q)$ in Definition~\ref{def-reward}, LLMs are also sensitive not only to \textit{which} facts are retrieved but also to \textit{how} they are presented. Specifically, empirical studies indicate that LLMs perform best with logically coherent and contextually organized fact sequences~\citep{chen2024premise,guo2025llms}, but suffer from fragmented or disordered inputs due to their inherent position bias and reasoning constraints~\citep{xiao2023efficient,jin2024long,yang2025ape}.
However,
graph-based retrievers are typically designed to be permutation-invariant for $\tau\in\cg$ and output unstructured sets of retrieval units (\textit{e.g.}, triples~\citep{li2024simple} or entities~\citep{mavromatis2024gnn}). As LLM evaluators 
$r(\cdot,q)$ exhibit strong sensitivity to structure and ordering, consequently, even a retrieved set $\wt$
with high coverage of $\wt^*$
may yield \textit{underspecified rewards} if the retrieved contents are poorly organized or misaligned with the LLM's reasoning preferences. Misaligned representation of retrieved information even exacerbates the complexity of identifying $\wt^*$
in the established optimization problem, and motivates
the need for structure-aware alignment in graph-based RAG.

\section{Weak-to-Strong Graph-based Retriever Alignment}
To align the weak retrievers to LLMs, we present \projfull (\proj), which strengthens the weak supervision and aligns the retrieved knowledge to the favorite forms of LLMs.

\subsection{Refining Supervision Signals with LLMs}\label{m1}

\begin{wrapfigure}[11]{r}{0.5\textwidth}  
\vspace{-0.15in}
  \centering
  \subfloat[\label{fig:2a}]{\includegraphics[clip,width=0.24\textwidth]{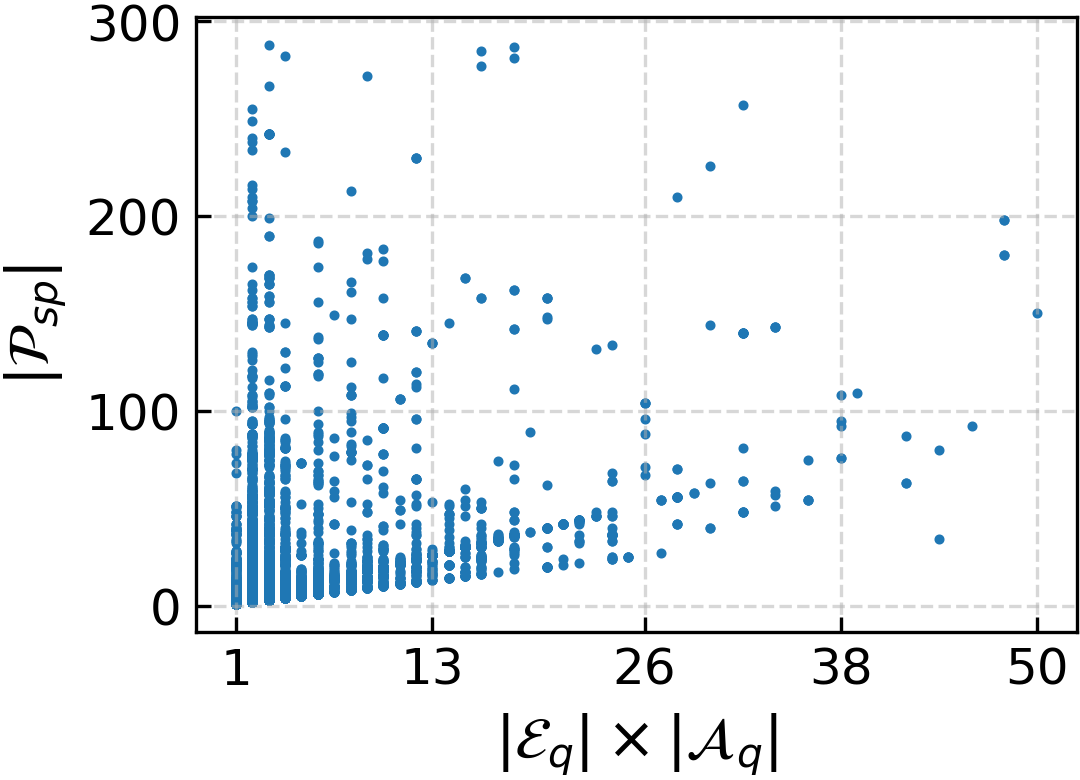}}
  \hspace{0.5em}
  \subfloat[\label{fig:2b}]{\includegraphics[clip,width=0.24\textwidth]{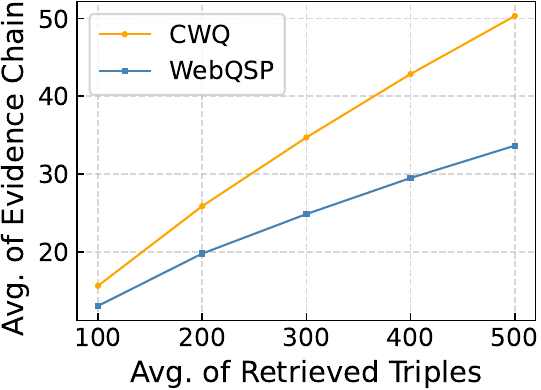}}
  \vspace{-0.5em}
  \caption{Scaling trends of: (a) $\cp_{sp}$ size versus the number of $q$-$a$ pairs, and (b) the number of generated reasoning chains versus retrieved triples.}
\end{wrapfigure}

To bridge the gap between $\wt^*$ and $\wt^w$ in a cost-efficient way, we first
construct a candidate path pool $\cp$, to cover diverse reasoning patterns, and then leverages LLMs to identify high-quality supervision signals from the candidates.

\textbf{Multi-Faceted Candidate Generation.}
Since shortest paths provide high-recall coverage~\citep{li2024simple}, we construct $\cp$ based on  $\cp_{sp}$, where each $P \in \cp_{sp}$ denotes a $q$-$a$ shortest path. 
To capture reasoning signals that shortest paths may overlook~\citep{gu2021beyond}, we incorporate two auxiliary subsets:
\begin{itemize}[leftmargin=*]
\vspace{-0.3em}
\setlength{\itemsep}{0pt}
\setlength{\parsep}{0pt}
\setlength{\parskip}{0pt}
\item $\cp_q$: \textit{Query-centric neighborhoods}, \textit{i.e.}, one-hop neighbors around each query entity, enabling LLMs to incorporate the direct properties of the query entities into their reasoning.
\item $\cp_a$: \textit{Answer-centric neighborhoods}, which enable comparisons across answer candidates based on numeric or categorical attributes.
\vspace{-0.3em}
\end{itemize}
Essentially, each candidate in $\cp_q$ and $\cp_a$ can be considered as a single-step reasoning path.
The final candidate pool is: $\cp:=\cp_{sp}\cup\cp_a\cup\cp_a$, providing a multi-faceted pool for LLM-guided refinement.

As shown in Fig.~\ref{fig:2a}, $\cp_{sp}$ can grow superlinearly with the number of $q$-$a$ pairs due to the redundancy in the candidate pool, which brings additional computational overhead. To optimize efficiency, we compress $\cp$ via structural merging to reduce redundancy. Details are given in Appendix~\ref{D.2}.

\textbf{LLM-Guided Candidate Refinement.} 
We leverage an LLM to identify plausible reasoning chains from the candidate pool $\cp$. 
Each candidate path $P\in\cp$ is textualized as a directed chain-of-entities that preserves logical flow for LLM reasoning.
Using in-context learning (ICL)~\citep{brown2020language} with explanation-based demonstrations, we prompt the LLM to identify a subset of candidates $\widehat{\cp}^+\subseteq\cp$ that provides sufficient and logically coherent evidence for answering the query. The detailed prompt can be found in Appendix~\ref{app:prompts}. Then, we extract the triples from $\widehat{\cp}^+$ as refined supervision signals $\wt^{+}$.

\begin{wraptable}{r}{0.35\textwidth}
\vspace{-0.15in}
\caption{Compression Effectiveness of the complexity control step. }
\vspace{-0.5em}
\label{tab:method-effective}
\begin{adjustbox}{width=\linewidth}
\begin{tabular}{lcc}
\toprule[1.5pt]
 GrailQA Dataset~\cite{gu2021beyond}          & \textbf{\# Avg} & \textbf{\# Max} \\ \midrule
Raw Candidates       & 275.16         & 98602          \\
+ Complexity Control & 13.54          & 137            \\
\textbf{Compression Ratio} & \multicolumn{1}{l}{$\sim$5\%} & \multicolumn{1}{l}{$\sim$0.14\%} \\ \bottomrule[1.5pt]
\end{tabular}%
\end{adjustbox}
\vspace{-0.3in}
\end{wraptable}

\textbf{Theoretical Discussion.}  
Essentially, the LLM-refined supervision signals $\wt^+$ provide a more accurate and semantically aligned approximation to $\wt^*$ than weak heuristics $\wt^w$. Empirically, as validated in experiments (Sec.~\ref{sec:exp}), the refined supervision demonstrates higher quality when applied to stronger training targets for the retriever. 

Meanwhile, \proj can overcome the computational bottleneck of the black-box optimization: as shown in Table~\ref{tab:method-effective}, our structural merging yields substantial compression (down to 5\% of the original candidate size on average), which is tractable for LLMs with standard reasoning capabilities under a constant order complexity $\mathcal{O}(1)$, despite the large $|\mathcal{G}|$.

\textbf{Retriever Training.} With the refined supervision, we can boost the retriever across different architectures (\textit{e.g.}, MLP, GNNs, or LLMs) and retrieval units (\textit{e.g.}, triple, entity, or path):
\begin{align}
    \max_\theta \mathbb{E}_{(q,\ca,\ct)\sim\cd}\left[\mathbb{P}_\theta(\wt^+\mid q,\ct)\right],\,\,\text{where}\,\,\wt^+:=\textsc{SignalRefiner}(q,\ca_q,\ct,\texttt{LLM})\label{eq:retrieval}
\end{align}
where $\mathbb{P}_\theta$ denotes the retriever distribution parameterized by $\theta$, and \textsc{SignalRefiner} represents LLM-based refiner. 
See Appendix~\ref{app:application} for instantiations of Eq.~\ref{eq:retrieval} across various retrieval granularities.

\subsection{Structure-Aware Post-retrieval Reorganization} \label{m3}

\begin{wrapfigure}{r}{0.35\textwidth}  
\vspace{-0.5in}
  \centering
\includegraphics[trim=110 10 0 10,clip,width=0.33\textwidth]{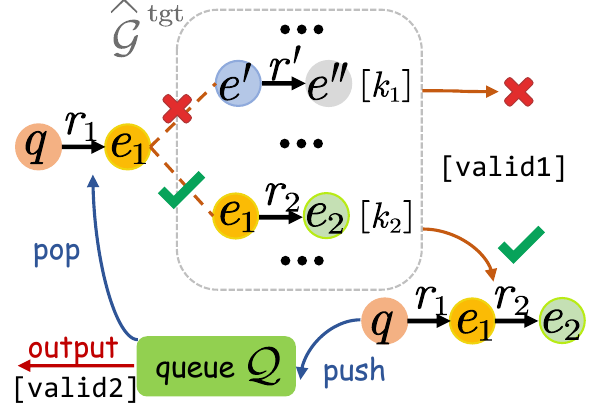}
  \caption{Illustration of BFS-guided chain expansion.}\label{fig:2d}
  \vspace{-0.4in}
\end{wrapfigure}

To bridge the representation gap between retrieval and reasoning (Sec.~\ref{challenge:2}), \proj also transforms the retrieved triples $\wt$ into a list of coherent evidence chains. Essentially, as the KG is originally organized in a logically coherent form, we align the retrieved results following the order in the KG.

\textbf{Chain Expansion.} 
Specifically, we perform BFS-guided chain expansion over $\wt$, which starts from query-anchored triples, defined as  $\wt^\text{src}:=\{(h,r,t)\in\wt\mid h\in\cE_q\lor t\in\cE_q\}$, and traverses through the left subgraph $\wt^\text{tgt} := \wt \setminus \wt^\text{src}$. Each chain is incrementally extended by appending entity-linked triples, where a new triple $(h_{k+1}, r_{k+1}, t_{k+1})$ can be appended only if its head entity $h_{k+1}$ matches the tail $t_k$ of the last triple in the current chain.
Alternatively, one could also apply other expansion methods, such as beam search. 

\textbf{Structure-Aware Scalability Control.} To further reduce redundancy and enhance semantic clarity, we introduce an additional structure-aware merging step that merges structurally related paths to improve scalability.  As shown in Fig.~\ref{fig:2b}, the number of the evidence chains produced through the two-step process grows approximately sub-linearly with the number of retrieved triples,
 indicating that our structure-aware reorganization strategy maintains
 tractable complexity. The chain expansion can be applied to various input units (\textit{e.g.}, triple and entity levels). More details of the aforementioned stages can be found in Appendix~\ref{app:details-sec-4.2}.

After the two steps, The query $q$ and reorganized retrieved evidence chains are integrated
into a structured prompt template with in-context demonstrations (\textit{c.f.}, Appendix~\ref{app:prompts}), guiding the LLM to generate factually grounded
answers.

\begin{table}[t]
\vspace{-0.15in}
\caption{Question-answering performance on WebQSP-sub and CWQ-sub. Best results are in bold. Avg. Rank reports the average rank of the evaluated methods across the four evaluation metrics. Best results are in bold.}
\label{tab:QA_main2}
\large
\resizebox{\textwidth}{!}{
\begin{tabular}{lcccccccccc}
\midrule[1.5pt]
                          & \multicolumn{5}{c}{Webqsp-Sub}                       & \multicolumn{5}{c}{CWQ-sub}  \\
                          \cmidrule(lr){2-6} \cmidrule(lr){7-11}
                          & Macro-F1 & Micro-F1       & Hit   & Hit@1 & Avg.Rank & Macro-F1       & Micro-F1 & Hit   & Hit@1          & Avg.Rank \\ \midrule
G-Retriever               & 54.13    & 23.84          & 74.52 & 67.56 & 11       & -              & -        & -     & -              & -        \\
RoG-Joint                 & 72.01    & 47.7           & 88.9  & 82.62 & 9        & 58.61          & 52.12    & 66.22 & 61.17          & 9.25     \\
RoG-Sep                   & 67.94    & 43.1           & 84.03 & 77.61 & 10       & 57.69          & 52.83    & 64.64 & 60.64          & 9.75     \\
SubgraphRAG (GPT-4o-mini) & 78.46    & 57.08          & 92.43 & 88.01 & 5.25     & 62.18          & 56.86    & 72.82 & 66.57          & 8        \\
SubgraphRAG (GPT-4o)      & 79.4     & 58.91          & 92.43 & 87.75 & 4.25     & 66.48          & 61.3     & 75.14 & 69.42          & 6.25     \\ \midrule
\projj@Triple (GPT-4o-mini) & 78.91          & 59.43 & \textbf{94.36} & 88.2           & 3   & 67.99 & 64.91         & 75.53          & 71.38 & 3.75 \\
\projj@Triple (GPT-4o)                    & \textbf{80.08} & 57.88 & 93.14          & \textbf{88.97} & 2.5 & 68.91 & \textbf{67.5} & \textbf{77.81} & 72.23 & 1.5  \\ \midrule
\projj@Entity (GPT-4o-mini)             & 77.84    & 57.03          & 93.78 & 86.79 & 5.75     & 66.74          & 62.63    & 75.95 & 70.65          & 4.5      \\
\projj@Entity (GPT-4o)                  & 79.03    & 56.92          & 92.5  & 88.77 & 4.5      & \textbf{69.62} & 66.35    & 76.16 & \textbf{73.14} & 1.5      \\ \midrule
\projj@Path (GPT-4o-mini)             & 78.44    & 60.55          & 91.4  & 85.76 & 6        & 65.01          & 61.77    & 73.24 & 66.92          & 6.75     \\
\projj@Path (GPT-4o)                  & 79.48    & \textbf{62.23} & 90.96 & 86.14 & 4.5      & 68.4           & 65.99    & 75.28 & 71.07          & 3.75     \\  \midrule[1.5pt]
\end{tabular}}
\vspace{-0.15in}
\end{table}
\section{Experiments}\label{sec:exp}

\subsection{Experimental Setup}
We conduct comprehensive  empirical studies to examine the effectiveness, efficiency, generalizability, and transferability of \proj in addressing the key  challenges in graph-based RAG (Sec.~\ref{Formulations}).  Specifically, we aim to address the following research questions: 
(\textbf{RQ1}) How effective is \proj in handling complex multi-hop reasoning tasks across various retrieval methods? 
(\textbf{RQ2}) If \proj indeed improves the supervision quality, can it also enhance data efficiency?
(\textbf{RQ3}) What is the contribution of each design component to the overall performance?
(\textbf{RQ4})
Can \proj also enhance state-of-the-art large reasoning models (LRMs)?
(\textbf{RQ5})
Are LLM-refined signals transferable across different backbone LLMs?
(\textbf{RQ6}) 
Do LLM-refined signals enable better out-of-distribution generalization in zero-shot settings?

\textbf{Datasets.} We adopt three prominent and challenging KGQA datasets that necessitate multi-hop reasoning -- WebQSP~\citep{yih2016value}, CWQ~\citep{talmor2018web}, and GrailQA~\citep{gu2021beyond}. All datasets utilize Freebase~\citep{bollacker2008freebase} as the underlying KG. For WebQSP and CWQ, we follow \citet{li2024simple} and establish WebQSP-sub and CWQ-sub, where we remove samples whose answer entities are \textit{absent} from the KG, to better evaluate the capability of LLM reasoners to produce factually consistent answers grounded in the external knowledge sources. The details of the evaluated datasets are given in the Appendix~\ref{app:datasets}.

\textbf{Baselines.} We mainly compare against several widely used  graph-based RAG methods, including
UniKGQA~\citep{jiang2022unikgqa},
KD-CoT~\citep{wang2023knowledge},
EtD~\citep{liu2024explore},
StructGPT~\citep{jiang2023structgpt},
ToG~\citep{sun2023think},
RoG~\citep{luo2023reasoning},
G-Retriever~\citep{he2024g},
SubgraphRAG~\citep{li2024simple},
GNN-RAG~\citep{mavromatis2024gnn}, and GraphRAG-FI~\citep{guo2025empowering}. More details are given in Appendix~\ref{app:baselines}.

\textbf{Metrics.} Besides the commonly reported Macro-F1 and Hit, we also include Micro-F1 to account for the imbalance in the number of answers across samples and Hit@1 for a more inclusive evaluation.

\begin{wraptable}{r}{0.45\textwidth}  
\vspace{-0.15in}
\caption{
Question-answering performance on WebQSP and CWQ. Best results are in bold. 
}
\vspace{-1em}
\label{tab:QA_main}
\centering
\begin{adjustbox}{width=\linewidth}
\begin{tabular}{lcccc}
\\
\toprule[1.5pt]
 & \multicolumn{2}{c}{WebQSP} & \multicolumn{2}{c}{CWQ} \\
\cmidrule(lr){2-3} \cmidrule(lr){4-5} 
      & Macro-F1     & Hit & Macro-F1 & Hit \\
\midrule
UniKGQA &
72.2 & 
- & 
49.0 &
- \\
KD-CoT  & 52.5 & 68.6 & - & 55.7 \\
ToG (GPT-4) & - & 82.6 & - & 67.6 \\
StructGPT & - & 74.69 & - & - \\
G-Retriever & 53.41 & 73.46 & - & - \\
RoG & 70.26 & 86.67 & 54.63 & 61.94 \\
EtD & - & 82.5 & - & 62.0  
\\
GNN-RAG &
71.3 &
85.7 &
59.4 &
66.8 \\
GraphRAG-FI &
75.98 &
91.89 &
60.34 &
71.12
 \\
SubgraphRAG  (Llama3.1-8B) & 70.57 & 86.61 & 47.16 & 56.98 \\
SubgraphRAG  (GPT-4o-mini) & 77.67 & 91.22 & 55.41 & 64.97 \\
SubgraphRAG  (GPT-4o) & 78.24 & 90.91 & 59.42 & 67.49 \\

\midrule\midrule
\multicolumn{5}{c}{\textbf{\projj@\textit{Triple}}} \\ \midrule
\proj  (Llama3.1-8B) & 69.91  & 87.39  & 51.24 & 65.02 \\
\proj  (GPT-4o-mini) &77.98 &	\textbf{92.87}  & 60.55 & 67.66 \\
\proj  (GPT-4o) & \textbf{78.76}	& 91.4  & {62.34} & \textbf{71.51} \\
\midrule\midrule
\multicolumn{5}{c}{\textbf{\projj@\textit{Entity}}} \\ \midrule
\proj  (Llama3.1-8B) & 65.59  &  86.0  & 49.73 & 61.8 \\
\proj  (GPT-4o-mini) & 76.78  & 92.26  & 58.13 & 66.24 \\
\proj  (GPT-4o) & 77.88    & 90.97  & \textbf{62.58} & 68.91 \\
\midrule\midrule
\multicolumn{5}{c}{\textbf{\projj\textit{@Path}}} \\ \midrule
\proj  (Llama3.1-8B) & 70.88  &	84.71  & 49.58 & 59.9  \\
\proj  (GPT-4o-mini) & 77.18  & 89.8  & 56.58 & 64.84 \\
\proj  (GPT-4o) & 78.4   & 89.56  & 60.79  & 67.23  \\
\bottomrule[1.5pt]
\end{tabular}
\end{adjustbox}
\vspace{-0.5in}
\end{wraptable}

\textbf{Implementation Details.}
(I) \textit{Retrieval.} We primarily use GPT-4o-mini and LLaMA-3.1-8B as backbone LLMs for supervision refinement, and select the retriever that achieves the best validation performance for the downstream stage. As \proj is agnostic to both retriever architectures and retrieval units, we instantiate it with three representative settings: (1) triple-level retrieval with an MLP-based retriever; (2) entity-level retrieval with a GNN-based retriever; and (3) path-level retrieval with an LLM-based retriever. We refer to these variants as \projj@Triple, \projj@Entity, and \projj@Path, respectively.
(II) \textit{Reasoning.} 
We perform zero-shot reasoning with different LLMs without any fine-tuning, including {LLaMA-3-8B}, {GPT-4o-mini}, GPT-4o, {QwQ-32B}, and {DeepSeek-R1}, with temperature set to 0 and random seed fixed to 42 for reproducibility.
More details of the implementations are given in Appendix~\ref{app:implementations}.

\subsection{Main Results on WebQSP and CWQ}\label{RQ12}

\textbf{Overall Performance (RQ1).} Tables~\ref{tab:QA_main} and \ref{tab:QA_main2} report evaluation results for \proj across three retrieval levels, \textit{i.e.}, Triple, Entity, and Path, paired with downstream reasoners of varying strength, including LLaMA-3-8B, GPT-4o-mini, and GPT-4o. \proj achieves state-of-the-art (SOTA) performance across all evaluated datasets, with all 12 metrics outperforming existing baselines when used with GPT-4o-mini or more advanced GPT-4o. Notably, \proj requires no finetuning of LLMs and invokes only one single reasoning call per query. Moreover, different retrieval levels of \proj excel on different metrics and datasets, reflecting its flexibility in accommodating varied reasoning demands and evaluation goals. 
For instance, the Path-level variant improves Micro-F1 by 5.6\% on WebQSP-Sub, while both the Triple and Entity levels perform strongly on CWQ (\textit{e.g.}, $\uparrow$10.1\% Micro-F1 on CWQ-Sub). 
Also, \proj scales gracefully with LLM capabilities, maintaining consistent performance gains from LLaMA-3-8B to GPT-4o, especially for CWQ \& CWQ-Sub datasets which requires deeper reasoning hops and handling multiple reasoning chains from different query entities.

\begin{figure}[t]
    \centering

    \begin{subfigure}[t]{0.48\textwidth}
        \centering
        \begin{minipage}[t]{0.49\textwidth}
            \includegraphics[width=\linewidth]{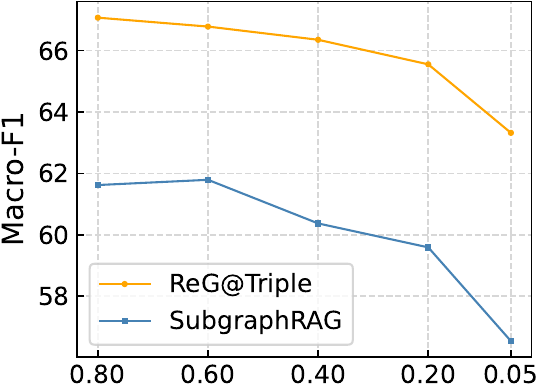}
        \end{minipage}
        \hfill
        \begin{minipage}[t]{0.49\textwidth}
            \includegraphics[width=\linewidth]{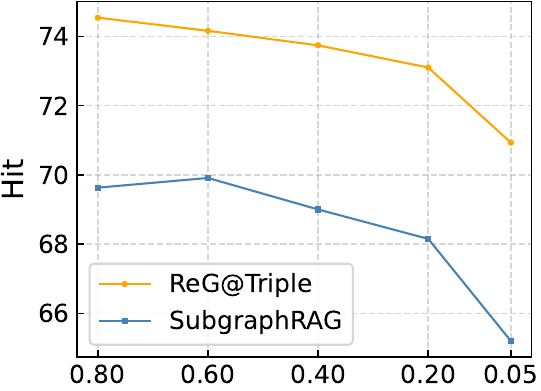}
        \end{minipage}
        \caption{Macro-F1/Hit -- Training ratio (80\%-5\%)}
    \label{fig:dy-a}
    \end{subfigure}
    \hfill
    \begin{subfigure}[t]{0.48\textwidth}
        \centering
        \begin{minipage}[t]{0.49\textwidth}
            \includegraphics[width=\linewidth]{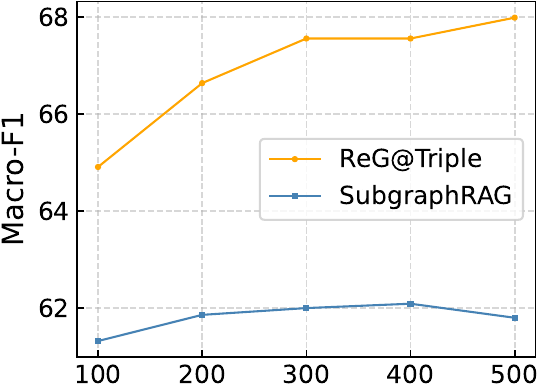}
        \end{minipage}
        \hfill
        \begin{minipage}[t]{0.49\textwidth}
            \includegraphics[width=\linewidth]{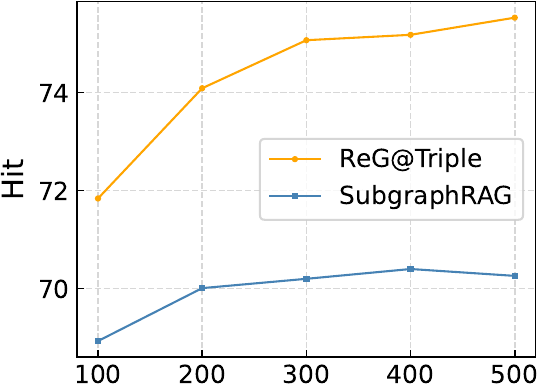}
        \end{minipage}
        \caption{Macro-F1/Hit -- Number of triples}\label{fig:dy-b}
    \end{subfigure}

    \caption{Performance comparison (Macro-F1 and Hit) between \projj@Triple and SubgraphRAG on CWQ-Sub: (a) versus training set ratio and (b) versus number of retrieved triples. Evaluated using GPT-4o-mini with identical retriever/reasoner configurations for fair comparison.}

\end{figure}

\textbf{Data Efficiency in Low-Resource Settings (RQ2).}
To further verify that \proj indeed refines the weak supervision, we examine the performance of \proj under the low-resource setting, where  only a small fraction of the full training dataset is available. 
As shown in Fig.~\ref{fig:dy-a}, we vary the proportion of training data from 80\% down to 5\%, comparing \projj@Triple with SubgraphRAG~\citep{li2024simple}, which is a state-of-the-art Triple-level baseline.  We observe that \proj with only 5\% training data achieves better  performance than the baseline counterpart with 80\% training data.
Moreover, SubgraphRAG suffers  more performance degradation as training data decreases, whereas \proj maintains more stable and stronger results. The robustness of \proj stems from the high-quality supervision signals refined via LLMs, which remove  noise and spurious signals and provide better guidance to the retriever.

Moreover,  since \proj requires LLM calls on only a small subset of the training dataset to yield a retriever that matches or outperforms fully supervised baselines, the overall computational cost associated with LLM refinement remains low. This underscores the data efficiency of our method and makes it practical for real-world deployment under limited supervision and computational budgets.

\textbf{Scalability with Retrieval Volume.}
It has been observed that the RAG performance may plateau or even degrade as retrieval volume increases~\citep{li2024simple}, due to the adverse effects of irrelevant content~\citep{wu2024easily} and the ``lost-in-the-middle'' phenomenon~\citep{liu2024lost}. 
As shown in Fig.~\ref{fig:dy-b}, SubgraphRAG exhibits limited ability to benefit from increasing retrieval content, even when equipped with strong GPT4o-mini. In contrast, \projj@Triple not only consistently outperforms SubgraphRAG but also shows more substantial performance gains as retrieval size increases, suggesting that \proj retrieved more accurate and semantically aligned information for LLM reasoning.

\subsection{Ablation Study (RQ3)}\label{RQ3}
\textbf{Effects of Individual Modeules}. 
We ablate the two key components of \proj: \textbf{S(I)} -- LLM-refined supervision signals, and \textbf{S(II)} -- structure-aware reorganization. 
Specifically, we consider two variants: one merely removes {S(I)}, denoted as ``w/o S(I)''; and the other removes both {S(I)} and {S(II)}, denoted as ``w/o S(I) \& S(II)'', to assess their individual and joint impacts across the three evaluated retrieval levels. Note that we do not evaluate ``w/o S(I) \& S(II)'' for the Path level as it takes reasoning paths as retrieval units.

As shown in Table~\ref{tab:ablation}, removing either component consistently degrades performance across all settings. The full model, which combines both \textbf{S(I)} and \textbf{S(II)} achieves the best results, highlighting the complementary roles of \textbf{S(I)} and \textbf{S(II)}: LLM-refined supervision improves retriever training quality, while structure-aware reorganization enhances the logical coherence of retrieved evidence. This is consistent with our discussion in Sec.~\ref{Formulations} on the benefits of tackling both challenges.

\begin{wrapfigure}{r}{0.5\textwidth}
    \vspace{-0.16in}
    \centering
    \captionsetup{aboveskip=4pt}

    \subfloat[\label{fig:decomp1}Multi-Entity (53.31\%)]{\includegraphics[width=0.23\textwidth]{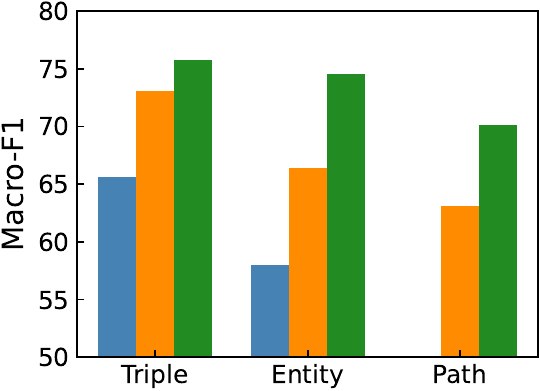}} \hspace{1em}
    \subfloat[\label{fig:decomp2}Multi-Hop (71.55\%) ]{\includegraphics[width=0.23\textwidth]{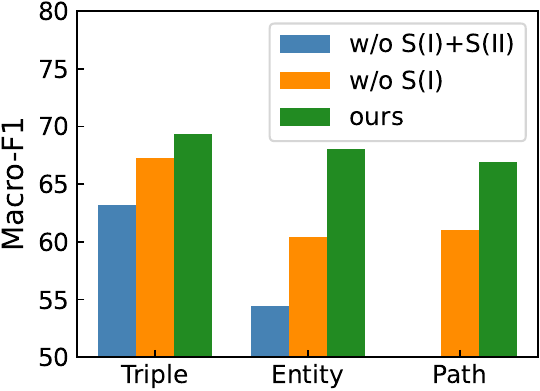}}
    \caption{Ablations across three retrieval levels over multi-hop and multi-entity queries on the CWQ-Sub dataset.Note that there are some overlapped samples between the two query types.  \proj \ consistently brings improvements with mutually beneficial components \textbf{S(I)} and \textbf{S(II)}. }
    \label{fig:decomp}
    \vspace{-0.19in}
\end{wrapfigure}
\textbf{Multi-Hop \& Multi-Entity QA}. Fig.~\ref{fig:decomp} reports ablation results on two challenging question types: multi-hop questions, where answers lie multiple hops away from the query entities, and multi-entity questions, which require reasoning over multiple query anchors. \proj shows notable gains on both types, indicating its strength in handling deep and compositional reasoning. The LLM-refined supervision proves especially effective for entity- and path-level retrieval, while structure-aware reorganization alone yields substantial improvements, particularly in multi-entity settings where aligning disparate reasoning chains is critical.
\clearpage

\begin{table}[t]
\caption{Ablation studies across triple, entity, and path levels on CWQ-Sub, under GPT4o-mini inference. S(I) and S(II) represent the two stages introduced in Sec.~\ref{m1} and \ref{m3}, respectively. }\label{tab:ablation}
\begin{adjustbox}{width=\linewidth}
\begin{tabular}{lcccccccccccc}
\toprule[1.5pt]
\multicolumn{1}{l}{} & \multicolumn{4}{c}{\textbf{Triple}} & \multicolumn{4}{c}{\textbf{Entity}} & \multicolumn{4}{c}{\textbf{Path}}  \\      \cmidrule(lr){2-5} \cmidrule(lr){6-9} \cmidrule(lr){10-13}                                                  
\multicolumn{1}{l}{} & Macro-F1 & Micro-F1 & Hit   & Hit@1 & Macro-F1 & Micro-F1 & Hit   & Hit@1 & Macro-F1             & Micro-F1             & Hit                  & Hit@1                \\ \midrule
\proj w/o S(I) \& S(II)         & 61.8     & 58.51    & 70.26 & 65.03 & 55.75    & 45.42    & 70.26 & 61.73 & - & - & - & - \\
\proj w/o S(I)               & 65.95    & 60.38    & 76.11 & 70.05 & 61.12    & 58.99    & 69.84 & 63.73 & 59.81                & 58.06                & 67.73                & 61.45                \\
\proj                 & 67.99    & 64.91    & 75.53 & 71.38 & 66.74    & 62.63    & 75.95 & 70.65 & 65.01                & 61.77                & 73.24                & 66.92                \\ \midrule[1.5pt]
\end{tabular}
\end{adjustbox}

\end{table}

\subsection{Effectiveness on State-of-the-Art Large Reasoning Models (RQ4)}\label{RQ4}

\begin{table}[t]

\caption{QA performance of \projj@Triple under QwQ-32B inference. See additional results on DeepSeek-R1 in Appendix~\ref{app:results}.}
\label{tab:LRMs}
\resizebox{\textwidth}{!}{%
\begin{tabular}{lccccccccc}
\toprule[1.5pt]
                   & \multicolumn{3}{c}{WebQSP-Sub} & \multicolumn{3}{c}{CWQ-Sub} & \multicolumn{3}{c}{GrailQA} \\\cmidrule(lr){2-4} \cmidrule(lr){5-7}  \cmidrule(lr){8-10}
                   & Macro-F1 & Hit & Avg.Tokens & Macro-F1  & Hit  & Avg.Tokens  & Macro-F1 & Hit & Avg.Tokens \\ \midrule
\proj w/o S(I) \& S(II) &  77.65        &  95.1   &      923.03      &   65.41        &  81.48    &   1351.14          &  78.73        &  89.85   &   886.13         \\ 

\proj w/o S(I)      &  76.51        &  94.55   &    702.82        &   66.38        &  80.2    &    971.2         &   81.26       &  90.3   &    710.4        \\
\proj              &  77.58        &  94.52   &  612.51          &    68.6      &  82.02    &    923.63         &  82.19        &  91.31   &   704.9         \\
\bottomrule[1.5pt]
\end{tabular}%
}
\vspace{-0.15in}
\end{table}

Following the advancement of LLMs, large reasoning models (LRMs) have gained huge success with significantly better capabilities in reasoning over the context~\citep{guo2025deepseek}.
To better understand the capabilities of LRMs in multi-hop reasoning and the effectiveness of \proj when paired with state-of-the-art large reasoning models, we apply \proj to {QwQ-32B} and {DeepSeek-R1}.

\textbf{Supervision Quality still Matters for LRMs.}
As shown in Table~\ref{tab:LRMs}, when paired with LRMs, retrievers trained with LLM-refined supervision signals still improve the performance by a noticeable margin. 
The improvements with more accurate and aligned retrievers reaffirm that the retrieval quality remains a crucial factor for highly capable LRMs.

\textbf{\proj Improves Reasoning Efficiency.}
Recently, it has been observed that LRMs often suffer from the \textit{overthinking problem}~\citep{chen2024not,sui2025stop}, which produces unnecessarily verbose reasoning traces with limited performance gains, ultimately degrading inference efficiency. 
To assess whether \proj alleviates this issue, we further examine the average reasoning tokens (Avg.Tokens) under different ablations. 

As shown in Table~\ref{tab:LRMs}, applying our structure-aware reorganization module leads to a substantial reduction in output length across all datasets. 
When further combined with refined supervision signals, the number of reasoning tokens can be reduced even further, accompanied by performance gains on various metrics, especially for datasets which require complex reasoning patterns.
Together, these results suggest that \proj can improve both reasoning efficiency and quality, even when used with highly capable LRMs, indicating its potential as a broadly applicable enhancement to existing QA pipelines.

\subsection{Transferability Analysis of LLM-Refined Signals (RQ5)}\label{RQ5}

\begin{wraptable}[13]{r}{0.45\textwidth}  
\vspace{-0.15in}
\caption{
Question-answering performance on WebQSP and CWQ when training retrievers over supervision signals refined by different LLMs, denoted as $\wt^+_\text{gpt}$ and $\wt^+_\text{llama}$. 
}
\vspace{-0.15in}
\label{tab:transfer-LLMs}
\centering
\begin{adjustbox}{width=\linewidth}
\begin{tabular}{ccccc}
\\
\toprule[1.5pt]
\multirow{2}{*}{\projj@Triple} & \multicolumn{2}{c}{WebQSP} & \multicolumn{2}{c}{CWQ} \\
\cmidrule(lr){2-3} \cmidrule(lr){4-5} 
      & Macro-F1     & Hit & Macro-F1 & Hit \\
\midrule

$\wt^w$ &76.62&92.14&57.55&64.91 \\\midrule
$\wt^+_\text{gpt}$ &77.03& 92.32&\textbf{60.55}&\textbf{67.66} \\
$\wt^+_\text{llama}$ &\textbf{77.98}&\textbf{92.87}&60.11&67.35 \\
$\wt^+_\text{gpt}\cap\wt^+_\text{llama}$ &77.51&92.75&59.38&66.47 \\
$\wt^+_\text{gpt}\cup\wt^+_\text{llama}$ &77.03&92.63&60.19&67.53 \\
\bottomrule[1.5pt]
\end{tabular}
\end{adjustbox}
\vspace{-4mm}
\end{wraptable}

When refining weak supervision signals, different LLMs may introduce distinct biases. To study the robustness and transferability of the refined signals, we ablate signals refined by GPT-4o-mini ($\wt^+_\text{gpt}$), LLaMA-3–8B ($\wt^+_\text{llama}$), their intersection ($\wt^+_\text{gpt}\cap\wt^+_\text{llama}$), and their union ($\wt^+_\text{gpt}\cup\wt^+_\text{llama}$), and evaluate QA performance via GPT-4o-mini inference, comparing them with the weak supervision signals ($\wt^w$).

As shown in Table~\ref{tab:transfer-LLMs}, we observe that the refined signals by different LLMs generically bring improvements compared to the original weak supervision $\wt^w$. 
Besides, the overall benefit of LLM refinement is more pronounced on CWQ than WebQSP. It is due to the more complex and compositional reasoning patterns in CWQ, which introduce more noise into weak heuristics-based supervision signals and increase the need for effective signal refinement.
Interestingly, neither the union nor the intersection of signals yields superior performance, indicating the need for more sophisticated strategies of collaborative refinement in the future.

\begin{table}[t]
\vspace{-0.15in}
\caption{Retriever generalizability to out-of-distribution (OOD) KGs and reasoning patterns. $A\to B$ denotes training retrievers on the training set $A$ and evaluating on dataset $B$.}
\label{tab:zeroshot}
\centering
\resizebox{0.85\textwidth}{!}{%
\begin{tabular}{lcccccccccc}
\toprule[1.5pt]
\multicolumn{1}{c}{} & \multicolumn{4}{c}{GrailQA $\to$ GrailQA-ZeroShot} & \multicolumn{4}{c}{CWQ $\to$ GrailQA-ZeroShot} \\ \cmidrule(lr){2-5} \cmidrule(lr){6-9}
\multicolumn{1}{c}{} &
  \multicolumn{1}{c}{Macro-F1} &
  \multicolumn{1}{c}{Micro-F1} &
  \multicolumn{1}{c}{Hit} &
  \multicolumn{1}{c}{Hit@1} &
  \multicolumn{1}{c}{Macro-F1} &
  \multicolumn{1}{c}{Micro-F1} &
  \multicolumn{1}{c}{Hit} &
  \multicolumn{1}{c}{Hit@1} \\\midrule
\proj w/o S(I) \& S(II)      & 75.59     & 44.02     & 84.02    & 81.67    & 67.29    & 35.16    & 76.1     & 73.79   \\
\proj w/o S(I)            & 80.73     & 50.13     & 87.06    & 84.35    & 71.07    & 40.74    & 78.48    & 75.88   \\
\proj         & 82.26     & 51.51     & 88.34    & 86.25    & 75.15    & 45.01    & 81.63    & 79.58  \\\toprule[1.5pt]
\end{tabular}

}\vspace{-0.15in}
\end{table}

\subsection{Out-of-Distribution Generalization in Zero-shot Settings (RQ6)}\label{RQ6}

We evaluate the out-of-distribution (OOD) generalization ability of retrievers trained with \proj under zero-shot settings. Specifically, we adopt the GrailQA dataset, which is explicitly designed to assess generalization to unseen schema items and domains. We first conduct few-shot training on the {GrailQA-Train} set and evaluate on {GrailQA-Dev-Zeroshot} subset. We then assess cross-dataset generalization by training on 20\% of {CWQ-Train} data and testing on the zero-shot subset.

As shown  in  Table~\ref{tab:zeroshot}, across both settings, retrievers trained with refined supervision consistently outperform those trained with weak proxy signals. These results suggest that LLM-refined supervision produces retrievers with stronger OOD generalization capabilities. This aligns with our earlier analysis in Sec.~\ref{challenge}, which highlights two key limitations of heuristic-based supervision: (1) it often includes {spurious paths} which reach the correct answer but cannot reflect the actual reasoning logic required to answer the query, and (2) it may {omit critical supporting information} that explains why an answer is correct.
In contrast, our LLM-guided refinement method results in more semantically faithful supervision and therefore yields stronger zero-shot generalization of retrievers.

\section{Conclusions}
In this work, we systematically studied the drawbacks of the weak retrievers in graph-based RAG. We showed that graph-based RAG can be formulated as a black-box combinatorial optimization problem. While resolving the original problem is computationally expensive due to the weak supervision and misaligned presentation of the retrieved results, we introduced \proj to align the weak retrievers to LLMs.
\proj first refines weak supervision signals from diverse candidate subgraphs guided by LLMs, and then structurally reorganizes retrieval outputs to better match the preferences of LLM reasoning. Extensive experiments across datasets, retriever types, and reasoning LLMs demonstrated that \proj consistently improves both retrieval accuracy, data efficiency, reasoning efficiency, and OOD generalizability, even when paired with state-of-the-art reasoning LLMs.

\clearpage

\bibliographystyle{references}
\nocite{}
\bibliography{references.bib}

\clearpage
\appendix
\clearpage
\section{Notation Summary}
\begin{table}[h]
\centering
\caption{Notations used in this work.}
\label{tab:notation}
\resizebox{\textwidth}{!}{
\begin{tabular}{cc}
\toprule[1.5pt]
\textbf{Symbol} & \textbf{Description} \\
\midrule
$\ct:=\{(h,r,t)\mid h,t\in\cE,r\in\cR\}$ & Original knowledge graph (KG), a set of triples $(h, r, t)$ \\
$\mathcal{E}_q$ & Set of query entities extracted from question $q$ \\
$\mathcal{A}_q$ & Set of answer entities associated with question $q$ \\
$P := (\tau_1, \dots, \tau_L)$ & Length-$L$ reasoning path where each $\tau_i = (h_i, r_i, t_i) \in \ct$ \\
$\mathbf{r}_P = (r_1, \dots, r_L)$ & Relation path corresponding to reasoning path $P$ \\
\midrule
$\wt$ & Retrieved subgraph from the KG (used for inference) \\
$\wt^\mathrm{src}:=\{(h,r,t)\in\wt\mid h\in\cE_q\lor t\in\cE_q\}$ & Query-anchored triples in the retrieved subgraph $\wt$ \\
$\wt^w$ & Weak supervision signals, \textit{i.e.}, a noisy subgraph used for training \\
$\wt^*$ & Oracle subgraph (ideal set of supporting facts for answering $q$) \\
\midrule
$\mathcal{P}:=\{P_i\}_{i=1}^{|\cp|}$ & Candidate path pool used for refinement \\
$\widehat{\mathcal{P}}^+$ & LLM-refined candidate paths selected from $\mathcal{P}$ \\
$\wt^+$ & LLM-refined supervision subgraph derived from $\widehat{\mathcal{P}}^+$ \\
\bottomrule[1.5pt]
\end{tabular}}
\end{table}

\section{Complementary Details of Formulations}\label{app:proof}
\subsection{Proof of Proposition~\ref{prop:general}}\label{proof:main}
\textit{proof.} 
Let $\wt^*\in\cH$ denotes the ground-truth hypothesis, where the hypothesis space satisfies $|\cH|=2^N$. Let $r^{(1:T)}:=(r^{(1)},\cdots,r^{(T)})$ denote the sequence of observed responses collected over 
$T$ rounds of interaction. From Fano's inequality,
\begin{align}
    H(\wt^*|r^{(1:T)})
    \leq \mathcal{O}(1)+\varepsilon \log|\cH|
\end{align}
From definitions of mutual information,
\begin{align}
    I(\wt^*;r^{(1:T)})=H(\wt^*)-H(\wt^*|r^{(1:T)}),
\end{align}
where $H(\wt^*)=\log|\cH|$, then we have,
\begin{align}
    I(\wt^*;r^{(1:T)})\geq (1-\varepsilon)\log|\cH| -\mathcal{O}(1) \label{fano}
\end{align}

From chain rules of mutual information,
\begin{align}
   I(\wt^*;r^{(1:T)})=\sum_{t=1}^TI(\wt^*;r^{(t)}|r^{(<t)})
\end{align}
Recall Eq.~\ref{reward}, since the $r(\cdot,q)$ depends on the number of correctly selected ($\sim\mathcal{O}(1)$) and incorrectly selected ($\sim\mathcal{O}(N)$) items,  the total number of possible reward values is at most $\mathcal{O}(N)$. Therefore, for the $t$-th term, we have,

\begin{align}
    { I(\wt^*;r^{(t)}|r^{(<t)})\leq H(r^{(t)}|r^{(<t)})\leq H(r^{(t)})}\leq\log|\cR|\leq \mathcal{O}(\log N),
\end{align}

where $\mathcal{R}:=\{r(\cs,q)\mid \cs\subseteq\ct\}$. THe first inequality holds via the property of mutual information. Hence,
\begin{align}
    I(\wt^*;r^{(1:T)})\leq T\mathcal{O}(\log N) \label{single-info}
\end{align}
From Eq.~\ref{single-info} and \ref{fano}, we have,
\begin{align}
    T\geq\Omega\left(\frac{(1-\varepsilon)N}{\log N}\right)
\end{align}
{If we fix the size $|\wt^{(i)}|$ in each round $i\in[T]$, then $|\cR|\leq |\wt^*|$ and hence $|\cR|\sim\mathcal{O}(1)$, and thus $T\geq\Omega((1-\varepsilon) N)$}

\subsection{A Specific Version}
Proposition~\ref{prop:general} provides a general lower bound on the number of queries $T$ for any algorithm interacting with the black-box evaluator.
We now analyze a specific iterative algorithm: Slightly different from Eq.~\ref{reward}, here we define the reward of a subset $\cs\subset\ct$ as,
\begin{align}
    r(\cs,q):=\frac{|\cs\cap\wt^*|s_0-|\cs\backslash\wt^*|\delta_0}{|\cs|s_0}\,\,\in[-\frac{\delta_0}{s_0},1],\label{reward-specific}
\end{align}
We define $r_0:=r(\ct,q)$, \textit{i.e.}, the reward on the universe set $\ct$, and let $r_0>0$. In each round of evaluation $i\in\{0,\cdots,T-1\}$, a set 
$\cs^{(i)}$ of size $S\ll N$ is drawn uniformly at random from $\ct$ and returned to the candidate pool after the round. Let the identified set $\wt^{(i)}$, initialized with $\wt^{(0)}=\oslash$, evolves via the following rule: 
\begin{align}
    \wt^{(i+1)}\leftarrow\wt^{(i)}\cup\cs^{(i)}, \,\,\text{if }\,\,r(\cs^{(i)},q)>\tau, \label{rule}
\end{align}
where the threshold $\tau\in(p,1)$ is a constant. Proposition~\ref{prop:specific} characterizes the sample complexity of recovering sparse high-relevance sets for the above algorithmic procedure.
\begin{proposition}
\label{prop:specific}
Guided by $r(\cdot,q)$ in Eq.~\ref{reward-specific} and rule~\ref{rule},  after T rounds, achieving
\begin{align}
    \mathbb{P}(\wt^*\subset\wt^{(T)})\geq 1-\varepsilon
\end{align}
with $\varepsilon\in(0,1)$ requires
\begin{align}
    T= \Omega\left(e^{2\tau^2S}\frac{N}{S}\log\frac{1}{\varepsilon}\right).
\end{align}
\end{proposition}
Specifically, stricter thresholds (higher $\tau$) exponentially increase the required rounds, as only subsets $\cs^{(i)}$ with higher signal-to-noise ratio meet the acceptance criterion $r(\cs^{(i)},q)>\tau$, and  thus  guarantees improvement in the Jaccard similarity between the final identified set $\wt^{(T)}$ and $\wt^*$.

\textit{proof.} We start by deriving $\mathbb{P}(t\notin\wt^{(T)})$ for a specific oracle item $t\in\wt^*$. We have,
\begin{align}
    \mathbb{P}(t\notin\wt^{(T)})=\prod_{i=0}^{T-1} \mathbb{P}(t\notin \wt^{(i+1)}|t\notin \wt^{(i)}) \label{decomp}
\end{align}
Now we focus on $\mathbb{P}(t\notin \wt^{(i+1)}|t\notin \wt^{(i)})$ for $i\in\{0,\dots,T-1\}$,
\begin{align}
    & \quad\,\,\mathbb{P}(t\notin \wt^{(i+1)}|t\notin \wt^{(i)})\\&=\mathbb{P}(t\notin \wt^{(i+1)}|t\notin \wt^{(i)},t\in\cs^{(i)})\mathbb{P}(t\in\cs^{(i)})+\mathbb{P}(t\notin \wt^{(i+1)}|t\notin \wt^{(i)},t\notin\cs^{(i)})\mathbb{P}(t\notin\cs^{(i)})\label{total_prob}
\end{align}
For $i$-th round, a subset $\cs^{(i)}$ of size $S$ is drawn at random from $\ct$ without replacement. Therefore, we have $\mathbb{P}(t\in\cs^{(i)})=\frac{S}{N}$ and $\mathbb{P}(t\notin\cs^{(i)})=1-\frac{S}{N}$. Also, it's easy to see $\mathbb{P}(t\notin \wt^{(i+1)}|t\notin \wt^{(i)},t\notin\cs^{(i)})=1$, and $\mathbb{P}(t\notin \wt^{(i+1)}|t\notin \wt^{(i)},t\in\cs^{(i)})=\mathbb{P}(r(\cs^{(i)},q)\leq\tau |t\in\cs^{(i)})$. 

Define $K:=|\wt^*|$. Let \textit{r.v.} $K_S$ be $|\cs^{(i)}\cap\wt^*|$, we know $K_S\sim\text{Hypergeometric}(N,K,S)$.
Via Eq.~\ref{reward-specific}, we have $K_Ss_0-(S-K_S)\delta_0\leq  S\tau s_0$, and,
\begin{align}
    K_S\leq S\frac{\tau s_0+\delta_0}{s_0+\delta_0}\triangleq S\theta,
\end{align}
where $\theta:=\frac{\tau s_0+\delta_0}{s_0+\delta_0}$.
Therefore,
\begin{align}
    \mathbb{P}(r(\cs^{(i)},q)\leq\tau |t\in\cs^{(i)})=\mathbb{P}(K_S\leq S\theta|t\in\cs^{(i)})\leq \mathbb{P}(K_S\leq S\theta),
\end{align}

and thus, we have,
\begin{align}
    \mathbb{P}(t\notin \wt^{(i+1)}|t\notin \wt^{(i)})\leq1-(1-\mathbb{P}(K_S\leq S\theta))\frac{S}{N}\triangleq 1-p_0\frac{S}{N}\label{single-final},
\end{align}
where $p_0:=\mathbb{P}(K_S\geq S\theta)$. 
Now, Eq.~\ref{single-final} $\to$ Eq.~\ref{decomp}, we have,
\begin{align}
    \mathbb{P}(t\notin\wt^{(T)})\leq\left(1-p_0\frac{S}{N}\right)^T
\end{align}
Applying the union bound over all $K$ target items gives
\begin{align}
    \mathbb{P}(\wt^*\not\subset\wt^{(T)})=\mathbb{P}\left(\bigcup_{t\in\wt^*}t\notin\wt^{(T)}\right)\leq\sum_{t\in\wt^*}\mathbb{P}(t\notin\wt^{(T)})\leq K\left(1-p_0\frac{S}{N}\right)^T\leq K\exp{\{-\frac{p_0ST}{N}\}}
\end{align}
To guarantee that this probability is at most $\varepsilon$, it suffices that
\begin{align}
    K\exp{\{-\frac{p_0ST}{N}\}}\leq \varepsilon,
\end{align}
which is equivalent to
\begin{align}
    T\geq\frac{N}{p_0S}\log\frac{K}{\varepsilon} \label{res_0}
\end{align}
Now, let us again focus on $p_0$. Define $p:=\be[K_S]=\frac{K}{N}$.
Firstly, we solve $\theta-p$. 
\begin{align}
    \theta-p&=\frac{\tau s_0+\delta_0}{s_0+\delta_0}-p
    =\frac{\tau s_0-(ps_0-(1-p)\delta_0)}{s_0+\delta_0}=\frac{\tau s_0-r_0}{s_0+\delta_0}
\end{align}
which holds via the definition of $r_0=r(\ct,q)=\frac{Ks_0-(N-K)\delta_0}{N}:=ps_0-(1-p)\delta_0$. 
Also, it's clear to see $r_0<ps_0<\tau s_0$ and thus $\theta >p$. Given that $r_0>0$, we have $\delta_0<\frac{p}{1-p}s_0$. Therefore, we have,
\begin{align}
    \theta-p>\frac{\tau s_0-ps_0}{s_0+\frac{p}{1-p}s_0}=(1-p)(\tau-p)
\end{align}
Via Hoeffding-type concentration inequality for $\text{Hypergeometric}(N,K,S)$ tails, we have
\begin{align}
    p_0=\mathbb{P}(K_S\geq S\theta)\leq\exp{\{-2S(\theta-p)^2\}},
\end{align}
and thus,
\begin{align}
    T\geq e^{2S(\theta-p)^2}\frac{N}{S}\log\frac{K}{\varepsilon}\geq  e^{2S(1-p)^2(\tau-p)^2}\frac{N}{S}\log\frac{K}{\varepsilon}
\end{align}
Considering $K\sim\mathcal{O}(1)$ and $p\sim\mathcal{O}(1/N)$, we have 
\begin{align}
    T= \Omega\left(e^{2\tau^2S}\frac{N}{S}\log\frac{1}{\varepsilon}\right)
\end{align}

\section{Details of Datasets and Baselines}
\subsection{Dataset Details}\label{app:datasets}
\textbf{WebQSP}~\citep{yih2016value} is an enriched version of the WebQuestions dataset, containing 4,737 questions annotated with semantic parses. The questions require up to 2-hop reasoning over the KG.

\textbf{CWQ}~\citep{talmor2018web} builds on WebQSP by extending questions with additional constraints or entity chains to form more complex multi-hop queries, totaling 34,689 questions with reasoning depths up to 4 hops. Notably,  over 50\% of WebQSP test questions (or their variants) appearing in CWQ's training set, and vice versa~\citep{li2024simple}. 

\textbf{GrailQA}~\citep{gu2021beyond} is a large-scale dataset designed to evaluate generalization in KGQA across i.i.d., compositional, and zero-shot settings. It features a wide range of logical forms and multi-hop questions. We focus on the zero-shot subset, which requires reasoning over unseen schemas and relation compositions, making it ideal for assessing retriever generalization to out-of-distribution knowledge and reasoning patterns.

\subsection{Baseline Details}\label{app:baselines}
\textbf{UniKGQA}~\citep{jiang2022unikgqa} unifies retrieval and reasoning for multi-hop QA over the KGs by integrating both stages in model architecture and parameter learning, tightly relating the retrieval and reasoning
processes.

\textbf{KD-CoT}~\citep{wang2023knowledge} enhances Chain-of-Thought reasoning in the LLMs for knowledge-intensive QA by verifying and refining intermediate steps through structured interaction with external knowledge, to reduce hallucinations and improve performance over standard CoT methods.

\textbf{EtD}~\citep{liu2024explore} is a two-stage framework for KGQA that combines lightweight GNN-based exploration with knowledge-enhanced prompting for LLM-based determination, enabling faithful reasoning over the KGs.

\textbf{StructGPT}~\citep{jiang2023structgpt}  introduces an Iterative Reading-then-Reasoning framework that equips the LLMs with specialized interfaces to iteratively gather evidence from structured data and perform focused reasoning.

\textbf{ToG}~\citep{sun2023think} introduces a training-free ``LLM $\otimes$ KG'' paradigm where an LLM agent iteratively explores related entities and relations on KGs via beam search to discover promising reasoning paths and perform traceable reasoning.

\textbf{RoG}~\citep{luo2023reasoning} proposes a planning-retrieval-reasoning framework which first generates relation paths grounded by KGs as faithful plans. These plans are then used to retrieve valid reasoning paths from the KGs for LLMs to conduct faithful reasoning.

\textbf{G-Retriever}~\citep{he2024g} perform RAG over textual graphs by formulating the task as a Prize-Collecting Steiner Tree optimization problem. It supports fine-tuning to enhance graph understanding via soft prompting.

\textbf{SubgraphRAG}~\citep{li2024simple} integrates a lightweight multilayer perceptron with a parallel triple-scoring mechanism for efficient and flexible subgraph retrieval while encoding directional structural distances to enhance retrieval effectiveness on capturing structural connections over KGs.

\textbf{GNN-RAG}~\citep{mavromatis2024gnn} combines GNNs with the language capabilities of LLMs in a RAG framework by retrieving candidate answers via GNN-based KG retrieval and guiding LLM inference using verbalized KG reasoning paths.

\textbf{GraphRAG-FI}~\citep{guo2025empowering} improves graph-based RAG by introducing a two-stage filtering mechanism and a logits-based selection strategy to reduce noise in retrieved information and balance external knowledge with LLMs' intrinsic reasoning.

\section{Methodology Details}\label{app:method-details}
\subsection{Examples of the Gap between the Oracle and Weak Supervision Signals}\label{app:eg-shortest}
Existing retrievers are typically trained on query-answer ($q$-$a$) shortest paths as proxy supervisions~\citep{zhang2022subgraph,luo2023reasoning}.
While easy to extract, these proxy signals face two intrinsic limitations:

(I) \textbf{Spurious inclusion ($\wt^w \setminus \wt^* \ne \varnothing$).} Shortest paths may contain relations that are semantically irrelevant to the query demand. For instance, consider the question ``Which countries border Spain'', 
the path of ``$\texttt{Spain} \xrightarrow{\texttt {currency-used}} \texttt{Euro} \xrightarrow{\texttt {country-used}} \texttt {Portugal}$'' fails to capture the spatial relation of interest (\textit{i.e.}, bordering), and thus provides misleading supervision. 

\textbf{(II) Incompleteness ($\wt^* \setminus \wt^w \ne \varnothing$).} Some queries require additional structural information beyond $q$-$a$ shortest paths. For \textit{aggregation}-type queries such as ``how many distilled spirits are associated with tequilla'', 
the correct answer is a numeric value derived from enumerating outgoing edges of the query entity. 
For \textit{comparison}-type queries such as ``Rocket engines with LOX oxidizer and chamber pressure < 79.0'', the $q$-$a$ shortest path ``$\texttt{Liquid oxygen} \xrightarrow{\texttt {rocket\_engines}} \texttt{Rocketdyne F-1}$'' provides a reasonable connection, but lacks critical evidence for the numerical constraint, which is captured by the answer-centric neighborhoods: ``$\texttt{Rocketdyne F-1} \xrightarrow{\texttt {chamber\_pressure}} \texttt{70.0}$'', enabling LLMs to perform necessary numerical filtering over candidate entity properties.

\subsection{Path Merging in LLM-Refinement}\label{D.2}
To preserve tractability and maintain minimal LLM usage overhead, we apply a structural merging step to compress the candidate path pool $\cp$ into a compact set, while retaining structural diversity and logical coverage.

(I)
\textbf{Answer Merging.} Only one representative answer $a^*\in\ca_q$ is selected from $\ca_q$, as correct answers typically share the same underlying reasoning path.

(II)
\textbf{Relation-Chain Merging.} Paths sharing the identical relation path are merged into a unified
candidate, removing redundancy without losing logical diversity. Here, the relation path of a path $P:=(\tau_1,\dots,\tau_L)$ is defined as the ordered sequence of relations $(r_1, \dots, r_k)$ extracted from each triple $\tau_i = (h_i, r_i, t_i)\in P$ in the path. 


\subsection{Instantiation of Retriever-Training Objectives }\label{app:application}
Eq.~\ref{eq:retrieval} gives a general-purpose MLE-based training objective which is model-agnostic and accommodates a wide range of retrieval units, including Triple, Entity, and Path. Here we give detailed instantiations of Eq.~\ref{eq:retrieval} across the three levels as follows.

\textbf{\projj@Triple}. Triples encapsulate atomic semantic relations between concepts and avoids the combinatorial explosion of subgraph enumeration.
We formulate retriever training as binary classification over individual triples $\tau \in \mathcal{G}$, where each triple serves as a basic retrieval unit. A triple is labeled as positive if $\tau \in \wt^+$, representing its semantic relevance to the query $q$.
Each triple is encoded in context: its representation $z_\tau := z_\tau(\tau, \mathcal{G}, q)$ incorporates both the local graph neighborhood and the query embedding, enabling the retriever to make relevance judgments that are both structure-aware and query-aware.
The retriever is optimized to maximize:
\begin{align}
\max_\theta\,\mathbb{E}_{\left(q, \mathcal{A}_q,\cg,\wt^+\right) \sim \mathcal{D}} \left[ \sum_{\tau \in \cg^+} \log p_\theta\left(\tau \mid z_\tau\right) + \sum_{\tau \in \mathcal{G} \setminus \cg^+} \log \left(1 - p_\theta\left(\tau \mid z_\tau\right)\right) \right],\label{obj-triple}
\end{align}
where $p_\theta$ denotes the retriever, and $z_\tau := z_\tau(\tau, \cg, q)$ denotes the contextualized representation of the triple $\tau$.

\textbf{\projj@Entity}. The entity-level retriever similarly treats individual entities $e \in \mathcal{E}$ as basic retrieval units. Entities that appear in $\wt^+$ are labeled as relevant, forming the positive set $\widehat{\cE}^+ := \mathcal{E}(\wt^+)$.
Each entity is encoded with a contextualized representation $z_e:=z_e(e,\cg,q)$ derived from its surrounding subgraph and the query. The model is trained to distinguish relevant entities from irrelevant ones via a binary cross-entropy loss:
\begin{align}
\max_\theta\, \mathbb{E}_{(q, \mathcal{A}_q, \mathcal{G}, \wt^+) \sim \mathcal{D}} \left[ \sum_{e \in \widehat{\cE}^+} \log p_\theta(e \mid z_e) + \sum_{e \in \mathcal{E} \setminus \widehat{\cE}^+} \log (1 - p_\theta(e \mid z_e)) \right].\label{obj-entity}
\end{align}

\textbf{\projj@Path}. Following \citet{luo2023reasoning}, we fine-tune a language model to serve as a planning module that generates plausible relation paths given a query $q$. These relation paths act as high-level logical sketches, which describes the reasoning patterns needed to traverse the KG and arrive at the correct answer. Once generated, these paths are grounded to the KG by matching them against actual sequences of triples, yielding reasoning paths that lead to candidate answers.
To train this planner, we use the relation paths extracted from each LLM-refined reasoning path $P \in \widehat{\cp}^+$ as supervision signals. The training objective maximizes the likelihood of generating the correct relation path given the query:
\begin{align}
\max_\theta\, \mathbb{E}_{(q, \mathcal{A}_q, \mathcal{G}, \widehat{\cp}^+) \sim \mathcal{D}} \left[\sum_{P \in \widehat{\cp}^+} \log p^\mathrm{LLM}_\theta(\mathbf{r}(P) \mid q)\right],\label{obj-relation}
\end{align}
where $\mathbf{r}(P)$ denotes the relation path of the reasoning path $P$ (see definition in Sec.~\ref{D.2}).

\subsection{Details of Two-Step Structure-aware Reorganization}\label{app:details-sec-4.2}

\subsubsection{BFS-based Evidence Chain Expansion}

We provide the full procedure of the BFS-based path expansion described in Sec.~\ref{m3}. Algorithm~\ref{alg:bfs_expansion} expands each query-anchored triple $\tau\in\wt^\mathrm{src}$ by iteratively growing coherent paths through entity-matching across triples in $\wt^\text{tgt}$. It's noteworthy that, for simplicity, the algorithm below only describes one expansion direction, \textit{i.e.}, extending the chain left-to-right, such that the query entity always appears as the head entity of the first triple. In practice, we perform bidirectional expansion: we also consider chains where the query entity serves as the tail of the final triple, expanding right-to-left accordingly.

\begin{algorithm}[t]
\caption{BFS-Based Reasoning Chain Expansion}
\label{alg:bfs_expansion}
\begin{algorithmic}[1]
\Require Retrieved triple set $\wt$, query entities $\mathcal{E}_q$, max length $L$
\Ensure Set of reasoning chains $\mathcal{P}$

\State $\wt^{\text{src}} \gets \{(h,r,t) \in \wt \mid h \in \mathcal{E}_q\}$ \Comment\texttt{Query-anchored triples}
\State $\wt^{\text{tgt}} \gets \wt \setminus \wt^{\text{src}}$ \Comment\texttt{Non-query triples}
\State $\mathcal{P} \gets \emptyset$, $\mathcal{Q} \gets \text{Queue}(\wt^{\text{src}}), \mathcal{T}^\mathrm{vis}\gets\emptyset$ \Comment\texttt{Initialize paths and queue}

\While{$\mathcal{Q}$ is not empty}
    \State $P \gets \mathcal{Q}.\text{dequeue}()$ \Comment\texttt{Current path $(\tau_1,...,\tau_k)$}
    \If{$P \notin \mathcal{P}$}
        \State $\mathcal{P} \gets \mathcal{P} \cup \{P\}$
        \State $(h_k,r_k,t_k) \gets \tau_k$ \Comment\texttt{Last triple in path}
        \State $(h_1,r_1,t_1) \gets \tau_1$ \Comment\texttt{First triple in path}
    \EndIf
    \If{$|P| \geq L$} \textbf{continue} \EndIf
    
    \For{each $\tau' = (h',r',t') \in \wt^{\text{tgt}}$}
        \If{$t_k == h'$ and $(h_1,t')\notin \mathcal{T}^\mathrm{vis}$} \Comment\texttt{Head-tail matching}
            \State $P' \gets P \oplus \tau'$ \Comment\texttt{Path extension}
            \State $Q.\text{enqueue}(P')$
            \State $\mathcal{T}^\mathrm{vis}\gets \mathcal{T}^\mathrm{vis}\cup \{(h_1,t')\}$
            \Comment\texttt{Mark query-target pair $(h_1,t')$ as visited}
        \EndIf
    \EndFor
\EndWhile
\State \Return $\mathcal{P}$
\end{algorithmic}
\end{algorithm}

While alternative heuristics such as beam search exist, we prefer BFS as its length limit $L$ directly corresponds to the number of reasoning hops, offering a more interpretable and semantically grounded control than its less meaningful counterparts such as the beam width $k$ in the beam search.
\subsubsection{Structure-aware Merging}
To further reduce redundancy and enhance semantic clarity, we merge structurally related paths using two operations:
\begin{itemize}[leftmargin=*]
    \vspace{-0.5em}
    \setlength{\itemsep}{0pt}
    \setlength{\parsep}{0pt}
    \setlength{\parskip}{0pt}
    \item[$\dagger$] \textit{Multi-answer merging.} Paths $P_1,\cdots,P_k$ are merged into a unified path if 1) rooted at the same query entity $e\in\ce_q$, 2) sharing the same relation path $(r_1,\cdots,r_{|P_1|})$, but 3) ending in different target entities.
    This handles scenarios where a single reasoning logic yields multiple valid answers and reduce the complexity of reasoning paths.
    \item[$\dagger$] \textit{Multi-entity merging.} After multi-answer merging, for multi-entity queries, \textit{i.e.}, $|\ce_q|>1$, paths $P_1$ and $P_2$ with distinct sources $\texttt{src}(P_1) \ne \texttt{src}(P_2) \in \mathcal{E}_q$ but overlapping targets $\texttt{tgt}(P_1) \cap \texttt{tgt}(P_2) \ne \emptyset$ are merged as follows:  
1) \textit{Spatial Reordering.} $P_1$ and $P_2$ are placed consecutively in the evidence sequence to reinforce their logical coherence;  
2) \textit{Target Refinement.} The new shared target set is set as $\texttt{tgt}(P_1) \cap \texttt{tgt}(P_2)$. 
While we describe pairwise merging here, the operation naturally generalizes to $n$-entity queries.

\end{itemize}

For the scalability of the proposed pipeline, as shown in Fig.~\ref{fig:2b}, the number of generated chains grows approximately sub-linearly with the number of retrieved triples,
 indicating that our structure-aware re-organization strategy maintains
 tractable complexity even under larger retrieval budgets.
 
After the two steps, The query $q$ and reorganized retrieved evidence chains are integrated
into a structured prompt template with in-context demonstrations (\textit{c.f.}, Appendix~\ref{app:prompts}), guiding the LLM to generate factually grounded
answers.

\section{Details of Experimental Implementation}\label{app:implementations} 
We adopt \texttt{gte-large-en-v1.5}~\citep{li2023towards} as the frozen text encoder for query and triple representation. This 434M model achieves a strong trade-off between efficiency and retrieval quality on English corpora, as validated by the Massive Text Embedding Benchmark (MTEB) leaderboard~\citep{muennighoff2022mteb}. In our main experiments, we evaluate three retrieval levels, \projj@Triple, \projj@Entity and \projj@Path. We introduce the detailed implementations as follows.

\textbf{\projj@Triple.}
We adopt the objective defined in Eq.~\ref{obj-triple}. For the retriever architecture, we follow \citep{li2024simple}, leveraging Directional Distance Encoding (DDE) to capture structural relationships in the KG. The encoded features are passed to a lightweight MLP for binary classification over individual triples. We rank all triples $\tau\in\cg$ by the retriever-output score $p_\theta(\tau)$ which quantifies its relevance to the query $q$, and the top-$K$ triples are selected to form the retrieved subgraph $\wt$ for downstream LLM reasoning. We set $K=500$ for GPT-4o and GPT-4o-mini, and $K=200$ for LLaMA-3.1-8B, considering its relatively limited reasoning capacity.

\textbf{\projj@Entity.}
We follow Eq.~\ref{obj-entity} as the training objective. The retriever is implemented as a GNN, specifically, a PNA (Principal Neighborhood Aggregation) network~\citep{corso2020principal}, which strengthens the expressive power of message passing by leveraging a rich combination of aggregators and degree-aware scalers. The retriever outputs a  score $p_\theta(e)$ for each entity $e\in\cE$. To maintain consistency with the \projj@Triple pipeline, we convert entity scores into triple scores by $s(\tau) := p_\theta(h) + p_\theta(t)$ for each $\tau = (h, r, t) \in \cg$. Top-$K$ triples are then selected for LLM inference.
A known limitation of this setup is that it ignores relation information, especially when multiple relations exist between a pair of entities. To mitigate this, we increase $K$ by 200 compared to the triple-level retriever and merge multiple parallel relations into a unified triple, \textit{e.g.}, $(h, r_1, t); (h, r_2, t) \rightarrow (h, r_1 \oplus r_2, t)$.

\textbf{\projj@Path.}
We follow Eq.~\ref{obj-relation} for supervision. Building on \citet{luo2023reasoning}, we fine-tune LLaMA-2-7B as a planning module to generate plausible relation paths given a query $q$. These predicted relation paths are then mapped onto the KG using beam search to retrieve grounded reasoning paths. We use a beam width of $k=20$ for GPT-4o and GPT-4o-mini, and $k=10$ for LLaMA3.1-8B to reflect their respective capacities.

\textbf{Other details.} Retriever training for the \projj@Triple and \projj@Entity variants, which adopt lightweight MLP and GNN (PNA) architectures respectively, is conducted on an RTX 4090 GPU. 
Both retrievers are trained for a fixed number of 80 epochs, and we select the checkpoint with the highest retrieval recall on the validation set for downstream evaluation. 
The LLM-based retriever used in \projj@Path relies on LLaMA-2-7B and is trained on two NVIDIA RTX A6000 GPUs using LoRA~\citep{hu2022lora} finetuning procedure on 8 epochs.

For the BFS-based chain expansion used in structure-aware reorganization, the maximum expansion length $L$ reflects the depth of reasoning allowed. We configure $L$ based on dataset characteristics: for the CWQ dataset, which requires complex multi-hop reasoning, we do not impose any upper limit on $L$; for all other datasets, we set $L = 2$ to balance path coverage and computational efficiency.

\section{Prompt Templates}\label{app:prompts}
\begin{figure}
\begin{tcolorbox}[colback=gray!5!white, colframe=gray!75!black, title=Input Prompts for LLM-guided Refinement]
\scriptsize

\textbf{System:}\\
\raggedright Given a question, an answer, and reasoning paths, select relevant paths for answering the question based on these criteria:

1. Relations in the selected path semantically align with certain question keywords;

2. Reject paths that contain only irrelevant details or demonstrate logical inconsistencies with question requirements.

Return all selected paths prefixed with `ans:', one per line.
\rule{\linewidth}{0.4pt} 
\\
\textbf{User:} \; \texttt{\color{red}{// ICL example}} \\
Paths:

{Path 0.} \\
boeing company $\rightarrow$ [spaceflight.rocket\_manufacturer.rockets\_manufactured] $\rightarrow$ saturn v rocket $\rightarrow$ [spaceflight.rocket.manufacturer] $\rightarrow$ North American Aviation \\

{Path 1.} \\
boeing company $\rightarrow$ [business.business\_operation.industry] $\rightarrow$ Aerospace $\rightarrow$ [business.industry.companies] $\rightarrow$ North American Aviation \\

{Path 2.} \\
Little Joe $\rightarrow$ [spaceflight.rocket.manufacturer] $\rightarrow$ North American Aviation \\

{Path 3.} \\
North American Aviation $\rightarrow$ [aviation.aircraft\_manufacturer.aircraft\_models\_made] $\rightarrow$ North American XB-70A \\

{Path 4.} \\
saturn v rocket $\rightarrow$ [spaceflight.rocket.height\_meters] $\rightarrow$ 110.6 \\

{Path 5.} \\
saturn v rocket $\rightarrow$ [spaceflight.rocket.country\_of\_origin] $\rightarrow$ United States of America \\

{Path 6.} \\
saturn v rocket $\rightarrow$ [spaceflight.rocket.mass] $\rightarrow$ 3038500.0 \\

{Path 7.} \\
Aerospace $\rightarrow$ [business.industry.companies] $\rightarrow$ North American Aviation

\vspace{1mm}

Question:

What rocket manufacturer produced Little Joe and also collaborated with Boeing on making a rocket with mass over 2.916e+06?
\\

Answer(s):

North American Aviation

\rule{\linewidth}{0.4pt} 
\\
\textbf{Assistant:} \; \texttt{\color{red}{// ICL example}} 

To identify the relevant reasoning paths, we need to focus on relationships that:

1) Confirm the manufacturer produced Little Joe, 2) Show collaboration with Boeing on a rocket, and 3) Provide information about the rocket mass.

Path 0 is valid because it establishes collaboration with Boeing (via Saturn V) and confirms North American Aviation as the manufacturer.

Path 2 is valid because it directly proves North American Aviation produced Little Joe.

Path 6 is valid because it provides information about the rocket mass (3,038,500 - 2.916e+06).

For irrelevant paths, Path 1/7 link Boeing and North American Aviation via Aerospace, which is indirect (no collaboration on making a rocket).

Path 3 discusses aircraft models, which is irrelevant to rockets.

Path 4/5 Describe background info of Saturn V (irrelevant to its mass).

Therefore, the relevant paths are:

ans: Path 0

ans: Path 2

ans: Path 6
\rule{\linewidth}{0.4pt} 
\textbf{User:}\; \texttt{\color{red}{// Sample to infer}} 

Paths:\\

\dots \;  \\
Question:\\
\dots \;  

Answers(s):\\
\dots \;  
\end{tcolorbox}
\vspace{-3mm}
\caption{Detailed prompt for LLM-based refinement used in our experiments.}
\label{fig:prompt_refine_detailed}
\end{figure}

\begin{figure}
\begin{tcolorbox}[colback=gray!5!white, colframe=gray!75!black, title=Input Prompts for Downstream QA]
\scriptsize

\textbf{System:}\\
\raggedright Answer the question using evidence chains from a knowledge graph. Return formatted answers, each on a new line and prefixed with `ans:'.
\rule{\linewidth}{0.4pt} 
\\
\textbf{User:} \; \texttt{\color{red}{// ICL example}} \\
{Evidence Chains}:

Chain 1.

Lou Seal → [sports.mascot.team] → San Francisco Giants → [sports.sports\_team.championships] → 2012 World Series; 2014 World Series; 2010 World Series

\vspace{1mm}

Question:

What year did the team with mascot named Lou Seal win the World Series?

\rule{\linewidth}{0.4pt} 
\\
\textbf{Assistant:} \; \texttt{\color{red}{// ICL example. Here the detailed explanation of the final answer(s) is optional.}} 

To find the year the team with mascot named Lou Seal won the World Series, we need to find the team with mascot named Lou Seal and then find the year they won the World Series.

From Chain 1, we can see that Lou Seal is the mascot of the San Francisco Giants, and that San Francisco Giants won the 2010, 2012, and 2014 World Series.

So, the team with mascot named Lou Seal (San Francisco Giants) won the World Series in 2010, 2012, and 2014.

Therefore, the formatted answers are:

ans: 2010 World Series

ans: 2012 World Series

ans: 2014 World Series
\rule{\linewidth}{0.4pt} 
\textbf{User:}\; \texttt{\color{red}{// Sample to infer}} 

Evidence Chains:\\

\dots \;  \\
Question:\\
\dots \;  

\end{tcolorbox}
\vspace{-3mm}
\caption{Detailed prompt for downstream QA used in our experiments.}
\label{fig:prompt_qa_detailed}
\end{figure}

Here we provide detailed prompt templates used in both the  LLM-guided supervision refinement stage (\textit{c.f.}, Fig.~\ref{fig:prompt_refine_detailed}) and the downstream QA stage (\textit{c.f.}, Fig.~\ref{fig:prompt_qa_detailed}). For both stages, we integrate a set of paths into the prompt. 
Specifically, for a $L$-length path $P:=(\tau_1,\cdots,\tau_L)$ where $\tau_i:=(h_i,r_i,t_i)$ denotes triples, we format it as a directional evidence chain of entities and their connecting relations, \textit{i.e.},  $h_1 \to$ \texttt{[}$r_1$\texttt{]} $\to t_1 \to$ \texttt{[}$r_2$\texttt{]} $\to t_2 \cdots\to t_L$ following a natural left-to-right progression for fluent LLM reasoning.  

For downstream QA prompts, our experiments show that the explanation component in the ICL example is optional for stronger LLMs. Particularly, we omit the explanation part in ICL when using GPT-4o and reasoning-focused LLMs, \textit{i.e.}, QwQ-32B and DeepSeek-R1, for downstream QA.

\section{Additional Experimental Results}\label{app:results}
We provide additional evaluations of QA performance of ReG@Triple under DeepSeek-R1 inference in Table~\ref{tab:add-dspk}.

\begin{table}[]
\caption{QA performance of \projj@Triple under DeepSeek-R1 inference. }
\label{tab:add-dspk}
\resizebox{\textwidth}{!}{%
\begin{tabular}{lcccccccccc}
\midrule[1.5pt]
                  & \multicolumn{5}{c}{WebQSP-Sub}                   & \multicolumn{5}{c}{CWQ-Sub}                      \\
                  \cmidrule(lr){2-6} \cmidrule(lr){7-11} 
                  & Macro-F1 & Micro-F1 & Hit   & Hit@1 & Avg.Tokens & Macro-F1 & Micro-F1 & Hit   & Hit@1 & Avg.Tokens \\ \midrule
w/o S(I) \& S(II) & 78.18    & 56.23    & 96.41 & 89.22 & 660.36     & 66.78    & 53.42    & 84.38 & 73.07 & 1026.53    \\
w/o S(I)          & 79.82    & 58.88    & 94.23 & 88.77 & 527.62     & 67.6     & 53.18    & 82.51 & 73.63 & 837.85     \\
\proj             & 80.32    & 59.13    & 93.71 & 87.68 & 531.92     & 69.87    & 53.67    & 84.22 & 74.44 & 799.119    \\ \midrule[1.5pt]
\end{tabular}%
}
\end{table}

\section{Future Work}\label{final}

Our pipeline is designed around the KGQA task and assumes the availability of entity-relation structured knowledge. Its generalization to other tasks or data modalities (\textit{e.g.}, vision-language) remains to be explored. The LLM-refined supervision signals may be not exactly the oracle one, and we primarily evaluate the effectiveness of such a LLM-refinement strategy  using automatic metrics on QA performance, lack of human evaluations due to the high cost. For the formulation part, incorporating LLM's inherent sensitivity to structure and position bias mentioned in Sec.~\ref{challenge:2} to the reward definition in Sec.~\ref{challenge} remains a future work.

\clearpage

\end{document}